\documentclass[review]{elsarticle}

\usepackage{hyperref}
\usepackage{multirow}
\usepackage{subfigure}
\usepackage[flushleft]{threeparttable}
\usepackage{booktabs}
\usepackage{comment}
\usepackage{amsmath}
\usepackage{algorithm}
\usepackage{algorithmicx}
\usepackage{algpseudocode}
\usepackage{xcolor}
\usepackage{amssymb} %RAFAEL ADICIONOU ESTE PACOTE
\journal{arXiv}

\begin{document}

\begin{frontmatter}

\title{Regularization Through Simultaneous Learning: A Case Study on Plant Classification}


\author[mymainaddress]{Pedro Henrique Nascimento Castro}
\ead{pedro.hnc@aluno.ufop.edu.br }

\author[secondaryaddress]{Gabriel Cássia Fortuna}
\ead{g.fortuna@unesp.br}

\author[mymainaddress]{Rafael Alves Bonfim de Queiroz}
\ead{rafael.queiroz@ufop.edu.br}

\author[mymainaddress]{Gladston Juliano Prates Moreira}
\ead{gladston@ufop.edu.br}

\author[mymainaddress]{Eduardo José da Silva Luz\corref{mycorrespondingauthor}}
\cortext[mycorrespondingauthor]{Corresponding author}
\ead{eduluz@ufop.edu.br}

\address[mymainaddress]{Computing Department, Federal University of Ouro Preto (UFOP), Zip Code: 35400-000, Ouro Preto, MG, Brazil}
\address[secondaryaddress]{Brazuca Lúpulos, Zip Code: 25610-080, Petrópolis, RJ, Brazil}

\begin{abstract}

In response to the prevalent challenge of overfitting in deep neural networks, this paper introduces Simultaneous Learning, a regularization approach drawing on principles of Transfer Learning and Multi-task Learning. We leverage auxiliary datasets with the target dataset, the UFOP-HVD, to facilitate simultaneous classification guided by a customized loss function featuring an inter-group penalty. This experimental configuration allows for a detailed examination of model performance across similar (PlantNet) and dissimilar (ImageNet) domains, thereby enriching the generalizability of Convolutional Neural Network models. Remarkably, our approach demonstrates superior performance over models without regularization and those applying dropout regularization exclusively, enhancing accuracy by 5 to 22 percentage points. Moreover, when combined with dropout, the proposed approach improves generalization, securing state-of-the-art results for the UFOP-HVD challenge. The method also showcases efficiency with significantly smaller sample sizes, suggesting its broad applicability across a spectrum of related tasks. In addition, an interpretability approach is deployed to evaluate feature quality by analyzing class feature correlations within the network's convolutional layers. The findings of this study provide deeper insights into the efficacy of Simultaneous Learning, particularly concerning its interaction with the auxiliary and target datasets.

\end{abstract}

\begin{keyword}
%\texttt{elsarticle.cls}\sep \LaTeX\sep Elsevier \sep template
%\MSC[2010] 00-01\sep  99-00
Regularization, Overfitting, Multi-task Learning, Simultaneous Learning, Hop classification.
\end{keyword}

\end{frontmatter}

%\linenumbers

\section{Introduction}

Neural networks have achieved significant results in various computer vision problems \cite{lecun2015deep, reedha2022transformer,makanapura2022classification,wan2022plant}. However, they often encounter a common issue: overfitting. Overfitting occurs when a model excels during training but performs substantially worse on test databases or real-world applications \cite{domingos2012few}. Several factors contribute to overfitting, including limited training data, model complexity, and data noise \cite{tian2022comprehensive}.

%
%A regularização é um dos métodos mais utilizados para diminuir o overfitting. Inicialmente, a palavra era atribuída aos termos de penalização adicionados à função de custo dos problemas de otimização. No entanto, atualmente, tem um sentido mais amplo e pode compreender qualquer modificação realizada em um algoritmo de aprendizagem para reduzir o erro na base de teste, e não no treinamento \cite{goodfellow2016deep}. Outra definição é que a regularização seja qualquer técnica que permita que um modelo generalize melhor \cite{kukavcka2017regularization}.
%
%Regularization is one of the most commonly used methods to reduce overfitting. Initially, the term referred to the penalty terms added to the cost function of optimization problems. However, currently, it has a broader meaning and can encompass any modification made to a learning algorithm to reduce error on the test set, rather than just during training \cite{goodfellow2016deep}. Another definition is that regularization is any technique that allows a model to generalize better \cite{kukavcka2017regularization}.

Regularization is among the most commonly used methods to reduce overfitting. Initially, the term referred to penalty terms added to the cost function of optimization problems. However, it now has a broader meaning and can encompass any modification made to a learning algorithm to reduce error on the test set rather than just during training \cite{goodfellow2016deep}. Another definition posits that regularization is any technique that enables a model to generalize better \cite{kukavcka2017regularization}.

Several regularization techniques exist, such as Weight Decay \cite{krogh1991simple}, Dropout \cite{srivastava2014dropout,warde2013empirical}, Batch Normalization \cite{ioffe2015batch}, Data Augmentation \cite{shorten2019survey}, Early Stopping \cite{zhang2005boosting}, Adversarial Training \cite{wong2020fast}, Transfer Learning \cite{pan2010survey}, and Multi-task Learning \cite{caruana1998multitask}. Although these techniques have successfully reduced overfitting, none can completely or universally solve this problem. Each of these techniques may produce different results when applied to problems of diverse nature \cite{moradi2020survey}. Therefore, combining two or more of these techniques is common to enhance a model's performance \cite{tian2022comprehensive}. Consequently, it is important to develop new techniques to achieve adequate generalization.

\begin{figure}[!ht]
\centering
%\hspace*{\fill}
\subfigure[Transfer learning]{
\includegraphics[width = .43\linewidth]{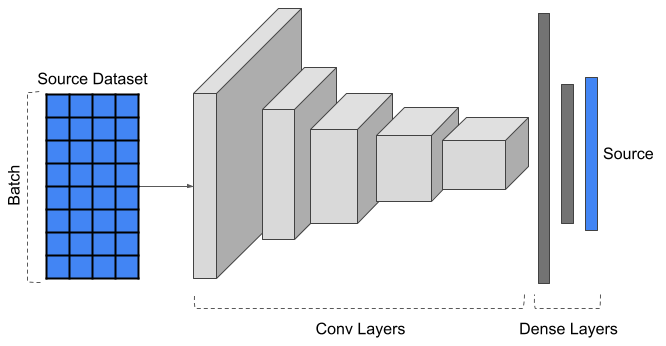}
\hspace{0.5cm}
\includegraphics[width = .43\linewidth]{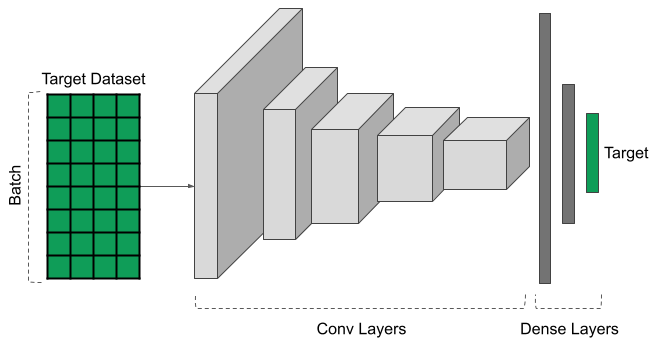}
\label{fig:model_transfer_learning}} 
\\
\subfigure[Multi-task learning]{
\includegraphics[width = .46\linewidth]{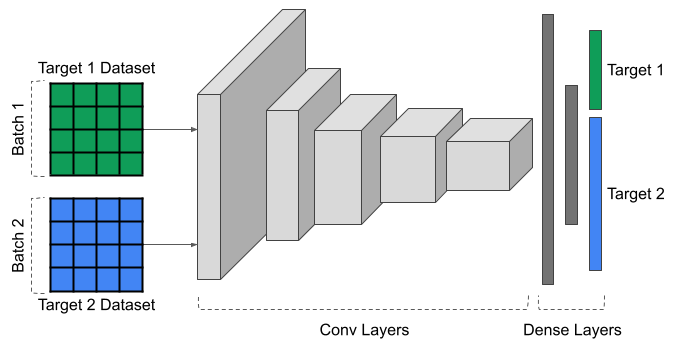}
\label{fig:model_multitask_learning}} 
%\\
\subfigure[Simultaneous Learning]{
\includegraphics[width = .49\linewidth]{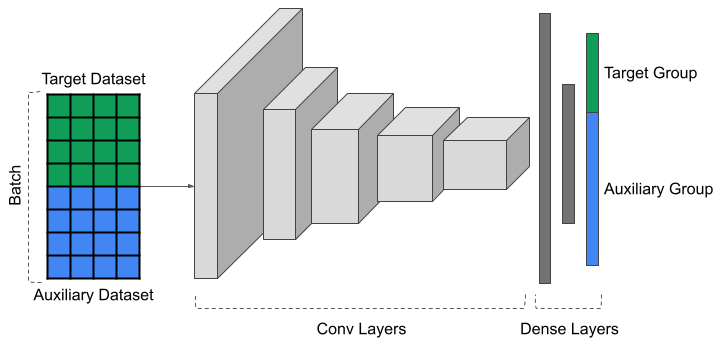}
\label{fig:model_simultaneous_learning}}
%\hspace*{\fill}
\caption{(a) An example of the Transfer Learning technique, where the last trained layer from a source dataset is replaced by a target dataset for fine-tuning. (b) An instance of Multitask Learning with two tasks, each with its respective input and output. (c) Simultaneous Learning technique, composed of a convolutional neural network with the last layer adapted to simultaneously classify images from the target group (in green) and the auxiliary group (in blue) during training.}
\label{fig:model}
\end{figure}

This study presents a regularization technique inspired by multi-task learning and transfer learning principles. Multi-task learning seeks to boost model performance by capitalizing on the intrinsic structure of related tasks \cite{ruder2017overview}, as illustrated in Figure \ref{fig:model_multitask_learning}, paralleling how humans apply prior knowledge to learning. Transfer learning, depicted in Figure \ref{fig:model_transfer_learning}, entails training a model on a source dataset before fine-tuning it on a target dataset. Transfer learning speeds up the training process of deep learning models and improves the model's ability to generalize \cite{abbas2023secure}. Combining multi-task learning and transfer learning yields a robust approach to reducing overfitting. Although multi-task learning has successfully addressed overfitting within the same domain \cite{huang2018automatic,wang2020environment,li2021multi}, training unrelated tasks concurrently may compromise performance, resulting in destructive interference or negative transfer \cite{kumar2012learning,zhao2018modulation,zhang2018overview,zhang2021survey}. This issue is particularly pronounced when data originate from different domains, which is common in deep learning due to the widespread adoption of transfer learning and the accessibility of pre-trained ImageNet models \cite{deng2009imagenet,russakovsky2015imagenet}. In light of the limitations and challenges highlighted in the paragraph above, an important research question arises: Is it possible to develop a composite loss function that effectively incorporates both target and auxiliary tasks, allowing for the simultaneous training of models using data from different domains? Investigating this question could lead to a more robust approach for addressing overfitting and improving model performance across various tasks and domains. In this study, we propose a method to address this issue within the context of a computer vision problem - the classification of hop varieties - by employing convolutional neural networks (CNNs) as depicted in Figure \ref{fig:model_simultaneous_learning}.

The approach entails modifying the model's final layer, the classification layer, to accommodate more outputs, thus enabling the simultaneous classification of two labeled datasets: the target dataset (See Figure \ref{fig:model_simultaneous_learning}, in green) and the auxiliary dataset (See Figure \ref{fig:model_simultaneous_learning}, in blue). As a result, the classes in the last layer are divided into two groups: the target group (in green) and the auxiliary group (in blue). This configuration allows the model to train on both datasets simultaneously, with the only alteration being the input batch of the network. To achieve this, Simultaneous Learning Loss is designed, assigning different weights to each group and introducing a penalty for incorrect classification between distinct groups. This approach aims to guide the model towards learning more general features from the target dataset rather than concentrating on specific aspects that may differentiate one class from another and potentially introduce noise. The objective is to optimize the model's overall performance and reduce the likelihood of overfitting the training data. It is crucial to note that the final layer, though comprising two groups, represents a single task, as the classification of each group is not independent, unlike Multi-task Learning. Once the training process is complete, the component responsible for classifying the auxiliary dataset can be discarded.

% For the purpose of experimental validation, a dataset for classifying hop species is explored. Hops, plants employed in beer brewing, possess varieties that directly impact taste, bitterness, and aroma \cite{oladokun2017perceived}. In certain countries, hops are sold with a certificate of origin that encompasses, among other attributes, the variety to which the hops belong \cite{mudura2010varietal}. Consequently, the precise identification of hop varieties is of utmost importance for hop growers \cite{salanctua2012determination}. Furthermore, the chosen dataset facilitates the examination of the proposed technique in unrelated domains and tasks since large datasets from distinct domains, such as ImageNet \cite{deng2009imagenet,russakovsky2015imagenet}, are readily available in the literature. Additionally, a dataset within a similar domain and task, such as PlantNet \cite{garcin2021plantnet}, can be investigated. Although our technique is focused on image classification, it can be expanded to other classification problems.

Plant classification is a subject of interest within the field of machine learning, facilitating the detection of weeds among crops \cite{burgos2010analysis}, the identification of vegetables with superior nutritional values \cite{hameed2018comprehensive}, and enabling non-professionals to easily discern between different species and varieties \cite{lee2017automatic}. In line with this growing interest, for empirical validation, we use a dataset centered on classifying plant species, specifically hops—a crucial ingredient in beer brewing. With over 250 cataloged variants of hops \cite{healey} exhibiting significant leaf similarity among varieties \cite{jenks2011plant}, this classification task represents a substantial challenge. Therefore, it's a relevant problem in the industry, demanding precision in identifying hop varieties. Beyond the application's relevance, our chosen dataset (UFOP-HVD) offers an opportunity to assess our proposed methodology's effectiveness across tasks within both similar and disparate domains. In particular, we draw upon PlantNet, which aligns with the UFOP-HVD's domain, and ImageNet, representative of a distinct domain. Importantly, both ImageNet \cite{deng2009imagenet,russakovsky2015imagenet} and PlantNet \cite{garcin2021plantnet} possess parallel characteristics—a large number of images and classes—thus rendering the UFOP-HVD dataset an ideal selection for our research.

Regarding model interpretability and explicability, we explore a technique termed Layer Correlation, which facilitates understanding the relationships between the features learned by the model. This method assesses the correlation level of the features generated for each class, determining whether the model can produce pertinent information for prediction. Moreover, we demonstrate the classes from the auxiliary group that most significantly activate the primary group and present their corresponding activation maps.

In the context of hop variety classification, with the aid of the method proposed herein, we have enhanced the state-of-the-art results previously presented in~\cite{castro2021end}. The contributions of this work can be summarized as follows:
\begin{itemize}
    %\item Uma metodologia de aprendizagem simultânea para regularizar um modelo;
    %\item Uma função de custo que penaliza erros inter-grupos;
    %\item Uma técnica de avaliação da qualidade das características geradas para comparação de modelos.
    \item A Simultaneous Learning method to regularize a model;
    \item A loss function that penalizes inter-group errors;
    %\item A technique to enhance the interpretation of the features learned in each layer of the model, thereby providing a deeper understanding of the learning process.
    \item A systematic approach to enhancing the interpretability of features learned at each model layer, promoting a deeper understanding of the learning process.
\end{itemize}
%
%A estrutura deste artigo é organizada da seguinte maneira: na Seção \ref{sec:related_works}, realizamos uma revisão dos trabalhos relacionados; na Seção \ref{sec:method}, apresentamos a função de custo proposta, a técnica de Simultaneous Learning, a métrica e a Layer Correlation; na Seção \ref{sec:experiments}, apresentamos as bases de dados utilizadas, as configurações e resultados dos experimentos e a interpretabilidade do modelo; por fim, na Seção \ref{sec:conclusion}, discutimos os resultados obtidos e propomos possíveis desdobramentos da pesquisa.

The structure of this paper is organized as follows: in Section \ref{sec:related_works}, we review related works; in Section \ref{sec:method}, we present the proposed cost function, Simultaneous Learning technique, evaluation metric, and Layer Correlation; in Section \ref{sec:experiments}, we describe the datasets used, the experimental configurations and results, and the model interpretability; finally, in Section \ref{sec:conclusion}, we discuss the results obtained and propose possible future research directions.

\section{Related Works}
\label{sec:related_works}
%
%\textbf{Multi-task Learning}. O Multi-task Learning tem como objetivo a aprendizagem simultânea de duas ou mais tarefas, possibilitando o compartilhamento de pesos entre elas. Essa abordagem visa aprimorar a generalização do modelo em tarefas correlatas, compartilhando informações durante o processo de aprendizado e permitindo que as tarefas se auxiliem mutuamente \cite{caruana1998multitask}. Tal método encontra inspiração na forma como os seres humanos utilizam conhecimentos prévios para aprender novas tarefas \cite{ruder2017overview}. Entretanto, o treinamento conjunto de tarefas não correlatas pode deteriorar a performance do modelo, gerando um efeito denominado interferência destrutiva ou transferência negativa \cite{kumar2012learning,zhao2018modulation,zhang2018overview,zhang2021survey}.

\textbf{Multi-task Learning}. The goal of Multi-task Learning is to perform the learning of two or more tasks simultaneously so that the learned weights can be shared among them. This aims to improve the model's generalization on related tasks, sharing information during learning and allowing tasks to assist each other \cite{caruana1998multitask}. This approach is inspired by how humans use prior knowledge to learn new tasks \cite{ruder2017overview}. However, the joint training of unrelated tasks may impair the model's performance, generating an effect called destructive interference or negative transfer \cite{kumar2012learning,zhao2018modulation,zhang2018overview,zhang2021survey}.
%
%Embora algumas pesquisas explorem tarefas não correlatas, como em \cite{paredes2012exploiting,liebel2018auxiliary,chennupati2019auxnet}, elas empregam uma única entrada com saídas distintas. Nesse cenário, não é viável utilizar bases de dados não relacionadas à base de dados alvo para o treinamento, uma vez que as tarefas, mesmo não sendo correlatas, são aplicadas somente aos mesmos dados. O presente estudo se distingue dos demais por permitir o uso de qualquer base de dados auxiliar, independentemente de sua relação com o domínio alvo, facilitando a busca por conjuntos de dados e expandindo as possibilidades de aplicação em diversas áreas. Adicionalmente, consideramos a classificação como uma única tarefa, possibilitando um melhor aproveitamento dos pesos aprendidos até as camadas finais, visto que as tarefas não são tratadas separadamente. Para mitigar o problema da transferência negativa, propomos uma penalização inter-grupos, que incentiva o modelo a diferenciar as características de cada grupo.

Although some research explores unrelated tasks, as in \cite{paredes2012exploiting,liebel2018auxiliary,chennupati2019auxnet}, they employ a single input with distinct outputs. In this scenario, it is not feasible to use databases unrelated to the target database for training, as tasks, even though not related, are applied only to data from the same domain. The present study stands out from others by allowing the use of any auxiliary database, regardless of its relation to the target domain, facilitating the search for datasets, and expanding the possibilities of application in various areas. In addition, we consider classification as a single task, which allows the learned weights to be better utilized until the final layers since the tasks are not treated separately. To mitigate the problem of negative transfer, we propose an inter-group penalty, which encourages the model to differentiate the features of each group.
%
%A seguir, apresentamos uma revisão de alguns dos estudos recentes que aplicam o aprendizado multi-tarefa na classificação de plantas.

Next, we review some recent studies that apply multi-task learning in plant classification.
%
%O estudo realizado por Zhu et al. \cite{zhu2019ta} tem como objetivo a classificação de espécies de plantas utilizando informações da família da planta e um mapa de calor das regiões mais importantes da planta. Primeiramente, uma rede neural convolucional com duas saídas é treinada, uma para a família e outra para a espécie, e um mapa de calor das regiões mais ativadas é gerado. Em seguida, a partir do mapa de calor, um recorte na imagem original da planta é gerado e este recorte é passado para uma segunda CNN que é treinada também com a família e espécie da planta. Os datasets utilizados foram o Malayakew, ICL, Flowers 102 and CFH plant.

The study conducted by Zhu \textit{et al.} \cite{zhu2019ta} aims to classify plant species using information about the plant family and a heatmap of the plant's most important regions. First, a convolutional neural network with two outputs is trained, one for the family and one for the species, and a heatmap of the most activated regions is generated. Then, using the heatmap, a crop in the original plant image is generated and passed to a second CNN, which is also trained with family and species information. The datasets used were Malayakew, ICL, Flowers 102, and CFH plant.
%
%Já a pesquisa de Lee et al. \cite{lee2021conditional} combina a classificação da espécie da planta com a identificação de doenças. Para isso, uma CNN é utilizada como modelo base, especificamente o InceptionV3, que compartilha os pesos para as duas tarefas. Entretanto, o resultado da classificação da espécie é introduzido antes da camada de classificação da doença com o objetivo de melhorar a acurácia. Um dataset próprio com 5334 imagens distribuídas em 311 espécies e 289 doenças foi construído para o trabalho.

The research by Lee \textit{et al.} \cite{lee2021conditional} combines plant species classification with disease identification. To do this, a CNN is used as the base model, specifically InceptionV3, which shares weights for both tasks. However, the result of the species classification is introduced before the disease classification layer in order to improve accuracy. A custom dataset with 5334 images distributed in 311 species and 289 diseases was constructed for the work.
%
%No trabalho de \cite{keceli2022deep}, os autores também abordam a classificação de espécies de plantas e a identificação de doenças. Para isso, utilizaram a arquitetura AlexNet como modelo base, que possui duas saídas softmax, uma para cada tarefa de classificação. A função de custo consiste na soma ponderada das funções de custo correspondentes a cada tarefa, ambas com peso igual a $0.5$. Os datasets utilizados foram o Plant Village e o FISB, que contêm amostras de arroz e milho.

In the work of \cite{keceli2022deep}, the authors also address the classification of plant species and disease identification. For this, they used the AlexNet architecture as the base model, which has two softmax outputs, one for each classification task. The cost function consists of a weighted sum of the corresponding cost functions for each task, both with a weight equal to $0.5$. The datasets used were Plant Village and FISB, which contain samples of rice and maize.
%
%O método proposto em \cite{wang2022dhbp} também envolve a classificação de plantas e o reconhecimento de doenças, utilizando a rede SE-ResNeXt-101 com saídas distintas para cada tarefa. A função de custo é a soma ponderada das duas saídas, em que a ponderação assume valores entre 0 e 1. Os datasets empregados foram o PlantVillage e o PlantDoc.

The method proposed in \cite{wang2022dhbp} also involves plant classification and disease recognition, using the SE-ResNeXt-101 network with different outputs for each task. The cost function is a weighted sum of the two outputs, where the weighting takes values between 0 and 1. The datasets used were PlantVillage and PlantDoc.
%
%\textbf{Transfer Learning.} A Transfer Learning (TL) é uma técnica que visa extrair conhecimento de uma tarefa fonte para uma tarefa alvo, ao invés de realizar a aprendizagem concomitante, como no Multi-task Learning \cite{pan2010survey}. Essa técnica geralmente consiste em treinar um modelo em uma grande base de dados rotulada (tarefa fonte) e, em seguida, realizar o fine-tuning em uma base de dados menor e específica (tarefa alvo), transferindo parte do conhecimento adquirido de uma base para outra. A TL é frequentemente utilizada para agilizar o treinamento e a convergência de modelos com muitos parâmetros, utilizando os pesos pré-treinados na base de dados maior. Nosso trabalho se enquadra na categoria Inductive Transfer Learning, onde os rótulos tanto do domínio alvo quanto do domínio auxiliar estão disponíveis \cite{zhuang2020comprehensive}.

\textbf{Transfer Learning.} Transfer Learning (TL) is a technique that aims to extract knowledge from a source task to a target task rather than performing concurrent learning, as in Multi-task Learning \cite{pan2010survey}. This technique usually involves training a model on a large labeled dataset (source task) and then fine-tuning it on a smaller, specific dataset (target task), transferring part of the acquired knowledge from one base to another. TL is often used to speed up the training and convergence of models with many parameters, using pre-trained weights on the larger dataset. Our work falls into the category of Inductive Transfer Learning, where both target and auxiliary domain labels are available \cite{zhuang2020comprehensive}.
%
%O estudo conduzido por \cite{kaya2019analysis} utilizou uma arquitetura de rede neural convolucional com três camadas, além das redes VGG-16 e AlexNet, para classificar espécies de plantas em quatro conjuntos de dados diferentes: Flavia, Swedish Leaf, UCI Leaf e Plantvillage. Para isso, foram empregados dois tipos de transferência de aprendizado. No primeiro método, os modelos foram pré-treinados com o conjunto de dados ImageNet e, posteriormente, realizou-se o fine-tuning em cada um dos conjuntos de dados de plantas. No segundo método, foram utilizados três dos conjuntos de dados de plantas para o pré-treinamento e o restante para o fine-tuning.

The study conducted by Kaya \textit{et al.} \cite{kaya2019analysis} used a convolutional neural network architecture with three layers, as well as VGG-16 and AlexNet networks, to classify plant species in four different datasets: Flavia, Swedish Leaf, UCI Leaf, and Plantvillage. For this, two types of transfer learning were employed. In the first method, the models were pre-trained on the ImageNet dataset and subsequently fine-tuned on each of the plant datasets. In the second method, three of the plant datasets were used for pre-training and the remaining for fine-tuning.
%
%Em \cite{espejo2020improving}, os autores investigaram os conjuntos de dados Plant Seedlings, que consiste em 12 espécies de cultivo, e Early Crop Weeds, que inclui 4 espécies de ervas daninhas. Foram testadas quatro arquiteturas de rede: VGG19, Inception-ResNet, Xception e Densenet. Para o processo de transferência de aprendizado, as redes foram pré-treinadas tanto com o ImageNet quanto com um conjunto de dados agrícola que os autores denominaram Agricultural Dataset. Esse conjunto de dados agrícola consiste em um dos conjuntos de dados de plantas (de cultivo ou de ervas daninhas), que é utilizado para o pré-treinamento, e o outro conjunto de dados é utilizado para o fine-tuning.

In \cite{espejo2020improving}, the authors investigated the Plant Seedlings dataset, consisting of 12 crop species, and the Early Crop Weeds dataset, which includes 4 weed species. Four network architectures were tested: VGG19, Inception-ResNet, Xception, and Densenet. The networks were pre-trained on both ImageNet and an agricultural dataset for the transfer learning process. This agricultural dataset consists of one of the plant datasets (either crop or weed), which is used for pre-training, and the other plant dataset is used for fine-tuning.
%
%O estudo realizado por \cite{espejo2020towards} apresenta um sistema de identificação de cultivo e erva daninha que combina diversas redes neurais convolucionais, incluindo Xception, Inception-Resnet, VGG16, VGG19, Mobilenet e Densenet, com classificadores de machine learning tradicionais, como Support Vector Machines, XGBoost e Logistic Regression. Todos os modelos foram pré-treinados com a base de dados ImageNet. Para essa pesquisa, foi elaborado um dataset próprio com imagens de tomate, algodão e duas espécies de ervas daninhas.

The study by \cite{espejo2020towards} presents a system for identifying crops and weeds that combines several convolutional neural networks, including Xception, Inception-Resnet, VGG16, VGG19, Mobilenet, and Densenet, with traditional machine learning classifiers such as Support Vector Machines, XGBoost, and Logistic Regression. All models were pre-trained with the ImageNet dataset. For this research, a custom dataset was created with images of tomato, cotton, and two weed species.
%
%O trabalho de \cite{ahmad2021performance} tem como objetivo detectar e classificar ervas daninhas em meio às plantações de milho e soja. A detecção foi realizada por meio da YOLOv3, enquanto as redes VGG16, ResNet50 e InceptionV3 foram utilizadas para a classificação. Todos os modelos foram pré-treinados com o ImageNet. O dataset utilizado na pesquisa foi criado a partir de imagens coletadas pelos autores e imagens adquiridas por meio de buscas no Google, totalizando 462 imagens distribuídas em quatro espécies de ervas daninhas.

The work of \cite{ahmad2021performance} aims to detect and classify weeds in corn and soybean plantations. Detection was performed using YOLOv3, while the VGG16, ResNet50, and InceptionV3 networks were used for classification. All models were pre-trained with ImageNet. The dataset used in the research was created from images collected by the authors and images acquired through Google searches, totaling 462 images distributed among four weed species.
%
%A abordagem proposta em \cite{pratondo2022classification} utilizou as redes VGG-19 e InceptionV3, pré-treinadas com o ImageNet, para classificar duas variedades de cúrcuma. O dataset utilizado na pesquisa foi construído a partir de fotografias tiradas com um smartphone em mercados, totalizando 647 imagens.

The approach proposed in \cite{pratondo2022classification} used the VGG-19 and InceptionV3 networks, pre-trained with ImageNet, to classify two varieties of Curcuma. The dataset used in the research was constructed from photographs taken with a smartphone in markets, totaling 647 images.
%
%Por fim, no estudo realizado por Chen et al. \cite{chen2022performance}, foram desenvolvidos métodos para identificação de ervas daninhas em plantações de algodão. Vários modelos foram avaliados, e o modelo ResNeXt101, pré-treinado com o conjunto de dados ImageNet, apresentou o melhor desempenho. Um conjunto de dados com 5187 imagens de 15 espécies de ervas daninhas foi criado e as imagens foram coletadas em condições de luz natural.

Finally, in the study by Chen \textit{et al.} \cite{chen2022performance}, methods were developed for weed identification in cotton plantations. Several models were evaluated, and the ResNeXt101 model, pre-trained with the ImageNet dataset, showed the best performance. A dataset with 5187 images of 15 weed species was created, and the images were collected under natural light conditions.
%
%Uma característica comum aos estudos mencionados é a substituição das camadas finais da rede para adaptá-la ao problema em questão. Durante o processo de fine-tuning, algumas camadas intermediárias podem ser congeladas, limitando a capacidade de aprendizagem do modelo, ou todas as camadas podem ser re-treinadas, resultando em esquecimento parcial do conhecimento adquirido no domínio alvo. O presente estudo se distingue da abordagem de Transfer Learning por descartar a classificação da base auxiliar somente após a conclusão total do treinamento e não antes do fine-tuning. Essa metodologia impede a perda de conhecimento adquirido da base de dados auxiliar durante todo o processo de aprendizagem.

A common characteristic of the mentioned studies is the replacement of the final layers of the network to adapt it to the specific problem. During the fine-tuning process, some intermediate layers can be frozen, limiting the learning capacity of the model, or all layers can be retrained, resulting in partial forgetting of the knowledge acquired in the target domain. The present study distinguishes itself from the Transfer Learning approach by discarding the classification of the auxiliary dataset only after the completion of the full training and not before the fine-tuning. This methodology prevents the loss of knowledge acquired from the auxiliary dataset throughout the entire learning process.
%
%\textbf{Dropout.} O Dropout é uma das técnicas de regularização mais utilizadas e bem-sucedidas na área de aprendizado profundo. Nós escolhemos esta técnica para comparar com o nosso método proposto. Durante o treinamento, em cada passo, alguns neurônios são aleatoriamente desligados, de modo que o modelo aprenda representações com menos capacidade de sobreajuste. Essa técnica ajuda a evitar o overfitting e é uma forma de simular um novo modelo mais simples a cada iteração. A versão padrão é utilizada como baseline para comparação de desempenho. No entanto, existem muitas variações, como o Dropconnect \cite{wan2013regularization}, Standout \cite{ba2013adaptive}, Curriculum dropout \cite{morerio2017curriculum}, DropMaps \cite{moradi2019sparsemaps}, Autodropout \cite{pham2021autodropout} e LocalDrop \cite{lu2021localdrop}.

\textbf{Dropout.} Dropout is one of the most used and successful regularization techniques in the deep learning area. We chose this technique to compare with our proposed method. During training, at each step, some neurons are randomly deactivated at each step so that the model learns representations with less overfitting capacity. This technique is a way to simulate a new, simpler model at each iteration. The standard version is used as a baseline for performance comparison. However, there are many variations, such as Dropconnect \cite{wan2013regularization}, Standout \cite{ba2013adaptive}, Curriculum dropout \cite{morerio2017curriculum}, DropMaps \cite{moradi2019sparsemaps}, Autodropout \cite{pham2021autodropout} and LocalDrop \cite{lu2021localdrop}.
%
%O dropout na versão padrão tem sido aplicado em diversos problemas de classificação de plantas, como identificação de flores \cite{liu2016flower}, classificação de plantas medicinais \cite{akter2020cnn}, reconhecimento de frutas \cite{wang2020fruit}, diferenciação entre cultivos e ervas daninhas \cite{haichen2020weeds} e identificação de mudas \cite{chauhan2021deep}. Alguns estudos sugerem que a combinação de diferentes técnicas de regularização pode ser eficiente \cite{van2017l2,he2019bag,li2019understanding}. Portanto, o dropout também será testado em conjunto com a nossa proposta para verificar se a combinação pode resultar em um melhor desempenho.

The standard version of dropout has been applied in several plant classification problems, such as flower identification \cite{liu2016flower}, medicinal plant classification \cite{akter2020cnn}, fruit recognition \cite{wang2020fruit}, differentiation between crops and weeds \cite{haichen2020weeds} and seedling identification \cite{chauhan2021deep}. Some studies suggest that combining different regularization techniques can be efficient \cite{van2017l2,he2019bag,li2019understanding}. Therefore, dropout will also be tested in conjunction with our proposal to verify if the combination can result in better performance.

\textbf{State of The Art For Hop variety Classification.} The method proposed in \cite{castro2021end} currently stands as the state-of-the-art solution for UFOP-HVD, facilitating a comprehensive, end-to-end process for hop variety classification. In this research, the authors conducted an exhaustive performance analysis of three eminent CNN architecture families: ResNets, EfficientNets, and InceptionNets. This investigation encompassed a series of ablation studies involving image classification, incorporating scenarios both with and without leaf segmentation, as well as with and without the application of data augmentation techniques.
Additionally, the authors proposed an ensemble architecture combining six distinct CNN models, a methodology termed Multi-cropped-FULL. This model utilizes multiple leaves from the same image in conjunction with the entire image as input, resulting in an accuracy rate of 81\%. Remarkably, with the integration of data augmentation techniques, the Multi-cropped-FULL model — considering all leaves of an image — reached an impressive accuracy of 95\%.
The authors also undertook a study termed ``cropped configuration'', wherein each detected leaf of the image (potentially comprising multiple hop leaves) was transformed into a new input for the problem, enhancing the number of images. In the case of ``cropped classification'', the employment of a ResNet50 architecture in conjunction with a 50\% dropout rate culminated in the most efficacious results, garnering an accuracy rate of 78\%.

\section{Methodology}
\label{sec:method}
%
%A abordagem de aprendizagem simultânea busca treinar um modelo para aprender representações de dois grupos, de forma a melhorar o desempenho em um deles, denominado grupo alvo, enquanto utiliza o outro grupo, chamado de grupo auxiliar, para auxiliar nesse processo. Este trabalho propõe uma regularização via aprendizagem simultânea, direcionada a modelos de deep learning, em particular, redes convolucionais de classificação. Neste contexto, a arquitetura das redes desempenha um papel fundamental. Nesta seção, serão apresentadas as funções de custo e as penalizações inter-grupos. Em seguida, serão exploradas as arquiteturas base e multi-grupo. Detalharemos as métricas utilizadas e definiremos a Layer Correlation.

The simultaneous learning approach aims to train a model to learn representations of two groups to improve performance on one of them, referred to as the target group, while using the other group, the auxiliary group, to assist in this process. This paper proposes a regularization via simultaneous learning for deep learning models, specifically convolutional neural networks. In this context, network architecture plays a crucial role. This section presents the loss functions and inter-group penalties. Then, the base and multi-group architectures are explored. The metrics used will be detailed, and Layer Correlation will be defined.

\subsection{Simultaneous Learning Loss}
%
%De maneira formal, considera-se que $f$ é um modelo que mapeia uma entrada $x$ para uma única saída contendo $n$ classes, que podem ser divididas em $s$ grupos ($G_1, G_2, ..., G_s$). Denotamos a saída do modelo para as classes do grupo $i$ como $[f(x)]_{G_i}$. Seja $L_i([f(x)]_{G_i})$ uma função de custo aplicada somente às saídas do modelo relativas ao grupo $i$. A função de custo geral, cujo objetivo é minimizar a soma ponderada das funções de custo de todos os grupos, pode ser definida como:
%
%Let $f$ be a model that maps an input $x$ to a single output containing $n$ classes, which can be divided into $s$ groups ($G_1, G_2, ..., G_s$). We denote the model's output for the classes in group $i$ as $[f(x)]_{G_i}$. Let $L_i([f(x)]_{G_i})$ be a loss function applied only to the model's outputs related to group $i$. The overall loss function, which aims to minimize the weighted sum of the loss functions for all groups, can be defined as:

Let $f$ be a model that maps an input $\mathbf{x}$ to a single output containing $n$ classes, which can be divided into $s$ groups ($G_1, G_2, ..., G_s$). Denote  the model weights by $\mathbf{\theta}$ and the model's output for the classes in group $i$ as $[f\left(\mathbf{x};\mathbf{\theta}\right)]_{G_i}$. Let $\mathcal{L}_i\left([f\left(\mathbf{x};\mathbf{\theta}\right)]_{G_i}\right)$ represent a loss function applied solely to the model's outputs related to group $i$. The overall loss function, aiming to minimize the weighted sum of the loss functions for all groups, can be defined as:
\begin{equation}
\theta^* = \arg\min_{\theta} \left\{ \mathcal{F}(\mathbf{\theta}, \Bar{\lambda}) = \sum_{i=1}^{s} \lambda_i \mathcal{L}_i\left([f\left(\mathbf{x};\mathbf{\theta}\right)]_{G_i}\right) \right \},
\label{eq:general_group_loss}
\end{equation}

\noindent wherein $\Bar{\lambda} = [\lambda_1, \lambda_2,\cdots, \lambda_s]$  has the parameters assigned to each loss function $i$ and $\theta^*$ denotes the model weights resulting from minimization.
%
%Para este trabalho em particular, limitamos a dois grupos, um principal, que chamamos de alvo, e um secundário, que chamamos de auxiliar. Dessa forma, a implementação da função de custo é dada por:
%
%For this particular work, we limit ourselves to two groups, a primary one, which we call the target, and a secondary one, which we call the auxiliary. In this way, the implementation of the cost function is given by:

In this work, we focus on two groups: a primary group, referred to as the target, and a secondary group, called the auxiliary. Consequently, the implementation of the loss function (Eq. \ref{eq:general_group_loss}) is expressed as:
\begin{equation}
\theta^* = \arg\min_{\theta} \left\{ 
 \mathcal{F}(\mathbf{\theta}, \Bar{\lambda}) = \lambda_t \mathcal{L}_t\left([f\left(\mathbf{x};\mathbf{\theta}\right)]_{G_t}\right) + \lambda_a \mathcal{L}_a\left([f\left(\mathbf{x};\mathbf{\theta}\right)]_{G_a}\right) \right\},
\label{eq:simultaneous_learning_group_loss}
\end{equation}
%
%em que $[f(x)]_{G_t}$ e $[f(x)]_{G_a}$ correspondem às saídas dos grupos alvo e auxiliar, respectivamente, $L_t$ e $L_a$ são suas funções de custo e $\lambda$ é a ponderação entre os dois grupos. Cabe ressaltar que, neste caso, apenas um hiperparâmetro precisa ser ajustado.
%
%where $[f(x)]_{G_t}$ and $[f(x)]_{G_a}$ correspond to the outputs of the target and auxiliary groups, respectively, $L_t$ and $L_a$ are their cost functions, and $\lambda$ is the weighting between the two groups. It should be noted that, in this case, only one hyperparameter needs to be adjusted.

\noindent where $[f\left(\mathbf{x};\mathbf{\theta}\right)]_{G_t}$ and $[f\left(\mathbf{x};\mathbf{\theta}\right)]_{G_a}$ correspond to the outputs of the target and auxiliary groups, respectively, $\mathcal{L}_t$ and $\mathcal{L}_a$ are their respective loss functions, 
$\Bar{\lambda} = [\lambda_t, \lambda_a]$. Let  $\lambda \in [0,1]$ a scalar be specified to the training of the model $f$. In this work, $\Bar{\lambda}$ is given by $\Bar{\lambda} = [\lambda, 1-\lambda]$ to depend on only one hyperparameter to be adjusted.
%
%No cenário mais simples, em que temos apenas o grupo alvo e nenhum grupo auxiliar, a função de custo se assemelha a um problema simples de classificação, onde $\lambda$ é igual a 1. Nesse caso, a categorical cross entropy \cite{bhatnagar2017classification,ho2019real}, apresentada na Eq. (\ref{eq:categorical_crossentropy}), pode ser adotada como a função $L_t$. A categorical cross entropy mede a divergência entre as distribuições de probabilidade das classes previstas pelo modelo e as classes reais das imagens de treinamento. Na equação, $\theta$ representa os parâmetros do modelo, $k$ é o número de classes do dataset alvo, $y$ é o vetor de rótulos verdadeiros, $\hat{y}$ é a saída do modelo com as predições e $i$ é o índice de cada classe. Segue a definição da equação:
%
%In the simplest scenario, where we have only the target group and no auxiliary group, the cost function resembles a straightforward classification problem, where $\lambda$ is equal to 1. In this case, the categorical cross entropy \cite{bhatnagar2017classification,ho2019real}, shown in Eq. (\ref{eq:categorical_crossentropy}), can be adopted as the $L_t$ function. The categorical cross entropy measures the divergence between the probability distributions of the classes predicted by the model and the true labels of the training images. In the equation, $\theta$ represents the model parameters, $k$ is the number of target dataset classes, $y$ is the ground-truth vector, $\hat{y}$ is the model's predictions, and $i$ is the index of each class. The equation is defined as follows:

In the simplest scenario, where only the target group is present, and no auxiliary group is considered, the loss function resembles a standard classification problem, with $\lambda$ equal to 1. In this case, the categorical cross entropy \cite{bhatnagar2017classification,ho2019real}, depicted in Eq. (\ref{eq:categorical_crossentropy}), can be employed as the $L_t$ function. Categorical Cross Entropy (CCE) quantifies the divergence between the probability distributions of the classes predicted by the model and the true labels of the training images. In the equation, $k$ represents the number of target dataset classes, $\mathbf{y} = [y_1, y_2, \cdots, y_k]$ is the ground-truth vector, $\hat{\mathbf{y}} = [\hat{y}_1, \hat{y}_2, \cdots, \hat{y}_k]$ signifies the model's predictions and $t$ is the index of each class. The equation is defined as follows:
\begin{equation}
	CCE(\mathbf{y},\hat{\mathbf{y}}) = -\sum_{t=1}^{k} y_{t} \log (\hat{y}_{t}).
	\label{eq:categorical_crossentropy}
\end{equation}
%
%Neste trabalho, propomos uma variação da função de custo categorical cross entropy, denominada Weighted Group Categorical Crossentropy, para o Simultaneous Learning. A equação correspondente é dada por:
%
%In this work, we propose a variation of the categorical cross-entropy loss function, called Weighted Group Categorical Crossentropy, for Simultaneous Learning. The corresponding equation is given by:

In this work, we propose a variant of the CCE loss function, termed Weighted Group Categorical Crossentropy (WGCC), designed explicitly for Simultaneous Learning. The corresponding equation is provided as follows:
\begin{equation}    
	WGCC(\mathbf{Y},\hat{\mathbf{Y}},\lambda) = - \lambda \sum_{t=1}^{k}  y_{t} \log (\hat{y}_{t})  
	- (1-\lambda) \sum_{a=k+1}^{k+m}  y_{a} \log (\hat{y}_{a}),\label{eq:weighted_group_categorical_crossentropy}
\end{equation}
\noindent where $\mathbf{Y} = [y_1, y_2, \cdots, y_k, \cdots,  y_{k+m}]$, $\hat{\mathbf{Y}} = [\hat{y}_1, \hat{y}_2, \cdots, \hat{y}_k, \cdots, \hat{y}_{k+m}]$ and $m$ is the number of classes in the auxiliary dataset.

%Na equação, $\theta$ representa os parâmetros do modelo, $k$ é o número de classes do dataset alvo, $m$ o número de classes do dataset auxiliar, $y$ é o vetor com os rótulos verdadeiros e $\hat{y}$ é a saída do modelo com as predições. O primeiro termo corresponde à saída do modelo referente à base de dados alvo, enquanto o segundo se refere à base de dados auxiliar. Ambos os termos são ponderados pelo hiperparâmetro $\lambda$, que indica a preferência que o modelo deve dar para o aprendizado de uma das bases de dados durante o treinamento.

%In the equation, $\theta$ represents the model parameters, $k$ is the number of classes in the target dataset, $m$ is the number of classes in the auxiliary dataset, $y$ is the ground-truth vector, and $\hat{y}$ is the model's predictions. The first term corresponds to the model output related to the target dataset, while the second refers to the auxiliary dataset. Both terms are weighted by the hyperparameter $\lambda$, which indicates the preference the model should give to learning from one of the datasets during training.

In Eq. (\ref{eq:weighted_group_categorical_crossentropy}), the first term corresponds to the model output associated with the target dataset with $k$ classes, while the second refers to the auxiliary dataset with $m$ classes. Both terms are weighted by the hyperparameter $\lambda$, dictating the model's preference for learning from one of the datasets during training.

In addition to the WGCC, we introduce an inter-group penalty term as per Eq. (\ref{group_penalty}) to deter the model from erroneously classifying samples from the target group as belonging to the auxiliary group and vice versa. In this manner, even if the model misclassifies certain samples, it is urged to constrain its errors within each group. The penalty term is derived by calculating the cross-product between the sum of predictions for one group and the sum of the ground-truth vector for the other group. As a result, the product is maximized when all predictions are in one group, and the ground-truth value resides in the other. The hyperparameters $\alpha$ and $\beta$ regulate the penalty factor. If the aim is to penalize only errors made on instances of the target group, simply set $\beta = 0$ and $\alpha > 0$. The same principle applies to errors committed on instances of the auxiliary group, in this case, setting $\alpha = 0$ and $\beta > 0$. The equation of the Group Penalty (GP) factor is presented as follows:

\begin{equation}    
	GP(\mathbf{Y},\hat{\mathbf{Y}},\alpha,\beta) = \alpha  \sum_{t=1}^{k} y_{t}  \sum_{a=k+1}^{k+m} \hat{y}_{a}
	+ \beta  \sum_{t=1}^{k} \hat{y}_{t}  \sum_{a=k+1}^{k+m} y_{a}.	
	\label{group_penalty}
\end{equation}

%A função de custo final, denominada \textbf{Simultaneous Learning Loss}, é a soma da Weighted Group Categorical Crossentropy com a Group Penalty, representada por:

%The final loss function, referred to as the \textbf{Simultaneous Learning Loss}, is the sum of the Weighted Group Categorical Crossentropy and the Group Penalty, represented by:

The final loss function, referred to as the Simultaneous Learning Loss (SLL), is the sum of the WGCC and the GP, represented by the following equation:
\begin{align}    
	SLL(\mathbf{Y},\hat{\mathbf{Y}}, \mathbf{h}) = 
	&- \lambda \sum_{t=1}^{k}  y_{t} \log (\hat{y}_{t})  
	- (1-\lambda) \sum_{a=k+1}^{k+m}  y_{a} \log (\hat{y}_{a})
    \nonumber
    \\
    &+ \alpha  \sum_{t=1}^{k} y_{t} \sum_{a=k+1}^{k+m} \hat{y}_{a} 
	+ \beta \sum_{t=1}^{k} \hat{y}_{t} \sum_{a=k+1}^{k+m} y_{a},
 \label{eq:simultaneous_learning_loss}
\end{align}

\noindent where $\mathbf{h} = [\alpha, \beta, \lambda]$ represent the hyperparameters vector of the model $f$.

\subsection{Architecture}

%Neste estudo, apresentamos duas arquiteturas: o modelo base e o modelo multi-grupo. O primeiro foi utilizado como referência base de desempenho, enquanto o segundo é empregado na metodologia Simultaneous Learning em conjunto com a Simultaneous Learning Loss.

This study presents two architectures: the base model and the multi-group model. The former serves as a performance baseline, while the latter is employed in the Simultaneous Learning methodology in conjunction with the SSL given by Eq.~\eqref{eq:simultaneous_learning_loss}.

\begin{figure}[!ht]
\centering
\subfigure[Base Model]{
\includegraphics[width = .46\linewidth]{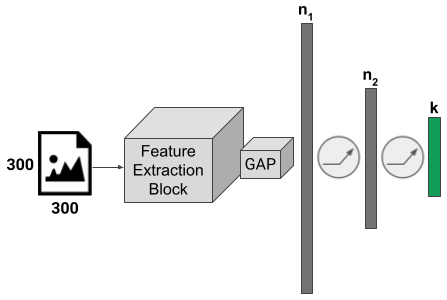}
\label{fig:model_base}} 
\hspace{2mm}
\subfigure[Multi-group Model]{
\includegraphics[width = .46\linewidth]{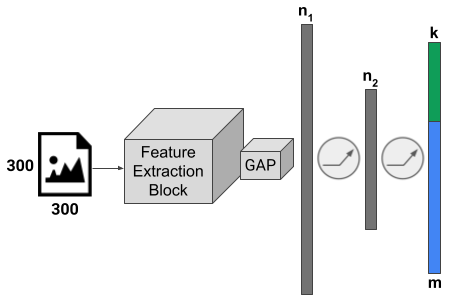}
\label{fig:model_multi_group}}
\caption{(a) Base model with a feature extraction block, a Global Average Pooling layer, and two dense layers, one with $n_1$ and another with $n_2$ neurons. The dense layers undergo a ReLU activation function. Ultimately, the model has a classification layer with $k$ outputs. (b) Multi-group model, a modification of the base model where $m$ outputs are added for the auxiliary group in the last layer, totaling $k + m$ outputs.}
\label{fig:model2}
\end{figure}

The base model's architecture comprises a series of convolutional layers responsible for extracting features from input images, which are resized to $300\times300$ pixels before processing. Following feature extraction, the output is channeled to a Global Average Pooling (GAP) layer \cite{lin2013network} that reduces the dimensionality of the features and enhances the model's robustness to variations in input image sizes. Subsequently, the GAP is connected to a sequence of two dense layers containing $n_1$ and $n_2$ neurons, respectively, utilizing the ReLU activation function. A final classification layer with $k$ outputs is added, where $k$ denotes the number of classes in the target dataset. The softmax activation function is applied to the classification layer outputs to determine each class's probability. The loss function employed in this model is categorical cross-entropy. Figure \ref{fig:model_base} visually represents the base model.

%A Figure \ref{fig:model_multi_group} exibe o modelo multi-grupo. A diferença estrutural para o modelo base é apenas na última camada, onde são acrescentadas $m$ saídas, sendo $m$ o número de classes do dataset auxiliar. Portanto, o total de saídas do modelo adaptado é $k + m$. O número de parâmetros treináveis a mais em relação à rede base é dado por $m \times (n_2 + 1)$, onde $n_2$ é o número de neurônios da penúltima camada do modelo. Pode-se observar que o crescimento do número de parâmetros é linear em relação às classes da base auxiliar, o que torna o tamanho e treinamento do modelo escaláveis. Ao final do treinamento, as $m$ saídas e seus respectivos parâmetros podem ser removidos, deixando apenas as saídas correspondentes ao dataset alvo.

%Figure \ref{fig:model_multi_group} illustrates the multi-group model. The structural difference from the base model lies only in the last layer, where $m$ outputs are added, with $m$ being the number of classes in the auxiliary dataset. Consequently, the total number of outputs for the adapted model is $k + m$. The additional trainable parameters compared to the base network are given by $m \times (n_2 + 1)$, where $n_2$ is the number of neurons in the penultimate layer of the model. It can be observed that the growth in the number of parameters is linear with respect to the classes of the auxiliary dataset, making the size and training of the model scalable. After training, the $m$ outputs and their corresponding parameters can be removed, leaving only the outputs related to the target dataset.

Figure \ref{fig:model_multi_group} depicts the multi-group model. The primary distinction from the base model resides in the last layer, where $m$ outputs are incorporated, with $m$ representing the number of classes in the auxiliary dataset. As a result, the adapted model's total number of outputs is $k + m$. Compared to the base network, the additional trainable parameters amount to $m \times (n_2 + 1)$, where $n_2$ denotes the number of neurons in the model's penultimate layer. Notably, the number of parameters increase is linear concerning the classes of the auxiliary dataset, rendering the model's size and training scalable. Upon completion of the training process, the $m$ outputs and their corresponding parameters can be discarded, retaining only the outputs pertinent to the target dataset.

\subsection{Training}

%O processo de treinamento, tanto para o modelo base, quanto para o modelo multi-grupo, é implementado utilizando o algoritmo de gradiente descendente para atualizar os parâmetros da rede neural. Enquanto o modelo base recebe apenas amostras do dataset alvo em cada lote de entrada, o multi-grupo recebe uma combinação de amostras do dataset alvo e auxiliar, sendo a proporção fixada em 50\% para cada conjunto de dados. É importante destacar que essa alteração na composição do lote não afeta a estrutura do modelo. No multi-grupo, durante cada época, todas as imagens de treinamento do dataset alvo são percorridas e, em cada passo, a outra metade do lote é completada com imagens aleatórias do dataset auxiliar, sem repetição por passo.

%The training process for both the base and multi-group models is implemented using the gradient descent algorithm to update the neural network parameters. While the base model only receives samples from the target dataset in each input batch, the multi-group model receives a combination of samples from the target and auxiliary datasets, with a fixed proportion of 50\% for each dataset. It is important to emphasize that this change in the batch composition does not affect the model structure. In the multi-group model, during each epoch, all training images from the target dataset are processed, and at each step, the other half of the batch is completed with random images from the auxiliary dataset, without repetition per step.

The training procedure for both the base and multi-group models utilizes the gradient descent algorithm to update the neural network parameters. In contrast to the base model, which solely receives samples from the target dataset in each input batch, the multi-group model obtains a combination of samples from both the target and auxiliary datasets, maintaining a fixed proportion of 50\% for each dataset. It is crucial to emphasize that this alteration in the batch composition does not impact the model's structure. The multi-group model processes all training images from the target dataset during each epoch. At the same time, at each step, the remaining half of the batch is supplemented with randomly selected images from the auxiliary dataset, ensuring no repetition per step.

%O pseudo-código para o processo de aprendizagem/treinamento do modelo multi-grupo é apresentado a seguir:

The multi-group model's computational strategy for the learning/training is systematized in Algorithm 1. In line 2 of this algorithm, the function \emph{length} returns the size of the target dataset's training set $\mathbb{T}$, i.e., the value $T$. In lines 6 and 8, it is used the notation $\mathbb{D}[I:J]$ to indicate that the images localized between the initial index $I$ and final index $J$ from the training set $\mathbb{D}$ under analysis are collected. In line 7, the function \emph{random} returns random images from the auxiliary dataset.

%usei a \usepackage{amssymb} %we added the package to the document

\begin{algorithm}
\caption{Pseudo-code for the multi-group model training}
\begin{algorithmic}[1]
\State \textbf{Input:} Pre-trained model $f$ with weights $\theta$, target dataset's training set $\mathbb{T} = \left[(x_{1}^t, y_{1}^t), (x_{2}^t, y_{2}^t), \cdots, (x_{T}^t, y_{T}^t)\right]$, auxiliary dataset's training set $\mathbb{A} =  \left[(x_{1}^a, y_{1}^a), (x_{2}^a, y_{2}^a), ..., (x_{A}^a, y_{A}^a)\right]$, hyperparameters vector $\mathbf{h} = [\alpha, \beta, \lambda]$, learning rate $\eta$, number of epochs $N$,  batch size $b$
\State $steps \gets length(\mathbb{T}) / b$ \Comment{Number of steps per epoch definition}
\State $b \gets b/2$ \Comment{Batch size division by half}
\For{$epoch \gets 1$ to $N$}
    \For{$s \gets 1$ to steps}
        \State $\mathbf{X}_{s}^t,\mathbf{Y}_{s}^t \gets \mathbb{T}[(s - 1)b:sb]$ \Comment{Target group images/labels}
        \State $\mathbb{A}^s \gets random(\mathbb{A})$ \Comment{Shuffling of samples in $\mathbb{A}$}      
        \State $\mathbf{X}_{s}^a,\mathbf{Y}_{s}^a \gets \mathbb{A}^s[1:b+1]$ \Comment{Auxiliary group images/labels}
        \State $\mathbf{X}_s,\mathbf{Y}_s \gets \mathbf{X}_{s}^t+\mathbf{X}_{s}^a,\mathbf{Y}_{s}^t+\mathbf{Y}_{s}^a$ \Comment{Images/labels concatenation}        
        \State $\hat{\mathbf{Y}}_s \gets f(\mathbf{X}_s,\mathbf{Y}_s, \theta)$ \Comment{Model $f$ prediction}       
        \State $\mathcal{C} \gets SLL(\mathbf{Y}_s, \hat{\mathbf{Y}_s}, \mathbf{h})$ \Comment{Evaluation of the cost function using Eq. \eqref{eq:simultaneous_learning_loss}}
        \State $\mathbf{g} \gets \nabla \mathcal{C}$ \Comment{Gradients with respect to model $f$ weights}
        \State $ \mathbf{\theta} \gets  \mathbf{\theta} - \eta \mathbf{g}$ \Comment{Model $f$ weights update}
    \EndFor
\EndFor
\State \textbf{Output:} Trained model $f$  with weights $\theta$.
\end{algorithmic}
\label{alg:ssl_training}
\end{algorithm}

\subsection{Metrics}

%A métrica de avaliação de desempenho utilizada neste trabalho é a acurácia delimitada (dacc), uma variação da acurácia comum. Durante o treinamento, as saídas da rede neural referentes à base de dados auxiliar não são removidas. Essa abordagem permite que, durante o teste na base de validação, apenas as saídas correspondentes à base de dados alvo sejam consideradas para determinar a classe, ignorando as saídas da base auxiliar. A acurácia delimitada tem outra vantagem, que é permitir que o modelo seja utilizado em produção sem remover as saídas da base auxiliar, possibilitando futuros retreinamentos.

%The performance evaluation metric used in this work is the delimited accuracy (dacc), a variation of the standard accuracy. During training, the neural network outputs related to the auxiliary dataset are not removed. This approach enables that, during testing on the validation dataset, only the outputs corresponding to the target dataset are considered for determining the class, disregarding the auxiliary dataset outputs. Delimited accuracy has another advantage, which is allowing the model to be used in production without removing the auxiliary dataset outputs, enabling future retraining.

In this study, we propose adapting the traditional accuracy metric, which we refer to as delimited accuracy (dacc). The dacc metric considers only some outputs (i.e., target group) from the neural network's classification results. In contrast, outputs associated with the auxiliary dataset are deliberately excluded during inference. This approach ensures that only the outputs corresponding to the target dataset are considered for class determination throughout the validation phase. Delimited accuracy offers an additional advantage: it allows the model to be deployed in production without removing the auxiliary dataset outputs, thereby facilitating potential future retraining.

\subsection{Interpretability}

%\textbf{Layer Correlation.} Para entendermos o modelo em um nível de features produzidas pelas camadas convolucionais, propomos um método de avaliação da correlação das features gerados por classe em cada camada. Neste método, cada camada é representada por um vetor composto pelas suas saídas.

\textbf{Layer Correlation.} To achieve a deeper understanding of the features generated by the model's convolutional layers, we put forth an approach that assesses the correlation of features produced for each class within each layer. Within this approach, each layer is represented by a vector comprising its outputs.

\begin{figure}[!ht]
\centering
\includegraphics[width = .96\linewidth]{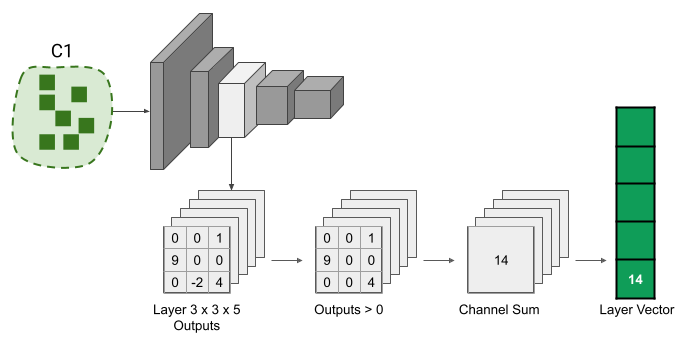} 
\caption{Example of transformation for class C1 of a layer with five channels of dimension $3 \times 3$ each into a five-element vector. Each channel has its positive values summed and becomes an element of the vector.}
\label{fig:layer_correlation_a}
\end{figure}

Initially, all images belonging to a single class from the test set of the target dataset are fed into the model for prediction. Subsequently, the positive outputs are summed for each channel within each layer. Ultimately, the summed value of the channel constitutes an element of a vector representing the layer. Figure \ref{fig:layer_correlation_a} demonstrates this process for class C1, utilizing the third layer of a model containing five channels, each with a dimension of $3x3$. The negative elements of the channels are excluded, followed by summing all elements of each channel, which yields a vector where each channel is represented as an element.

%Esse processo é repetido para cada classe, resultando em uma lista de vetores para cada uma delas, conforme exemplo na Figura \ref{fig:layer_correlation_b} para as classes C1, C2 e C3.

%This process is repeated for each class, resulting in a list of vectors for each of them, as shown in the example in Figure \ref{fig:layer_correlation_b} for classes C1, C2, and C3.

\begin{figure}[!ht]
\centering
\includegraphics[width = .36\linewidth]{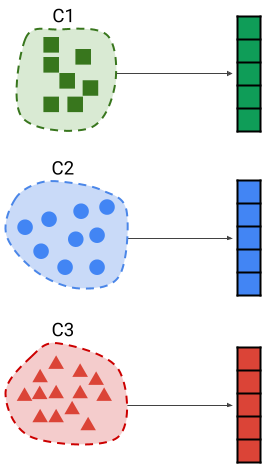} 
\caption{Example of vectors from a single layer of three different classes.}
\label{fig:layer_correlation_b}
\end{figure}

This procedure is repeated for every class, resulting in a list of vectors corresponding to each class, as demonstrated in the example in Figure \ref{fig:layer_correlation_b} for classes C1, C2, and C3.

%Em seguida, são obtidos os coeficientes de correlação de Pearson \cite{rodgers1988thirteen} entre os vetores de cada classe. Um escalar é gerado a partir da média do valor absoluto desses coeficientes. Quanto maior o valor, maior a correlação entre as features geradas pela camada em avaliação. A alta correlação pode levar a um pior desempenho do modelo, uma vez que variáveis dependentes, ou seja, altamente correlacionadas, fornecem pouca ou nenhuma informação adicional sobre as classes \cite{chandrashekar2014survey}.

%Next, Pearson's correlation coefficients \cite{rodgers1988thirteen} are obtained between the vectors of each class. A scalar is generated from the mean of the absolute value of these coefficients. The higher the value, the greater the correlation between the features generated by the layer under evaluation. High correlation can lead to worse model performance, as dependent variables, i.e., highly correlated ones, provide little or no additional information about the classes \cite{chandrashekar2014survey}.

Subsequently, Pearson's correlation coefficients \cite{rodgers1988thirteen} are calculated between the vectors of each class. A scalar is derived from the mean of the absolute values of these coefficients. A higher value signifies a stronger correlation between the features generated by the layer under evaluation. High correlation may contribute to diminished model performance, as dependent variables or highly correlated ones offer little to no supplementary information about the classes \cite{chandrashekar2014survey}.

\textbf{Grad-CAM.} In this study, we also employ Grad-CAM (Gradient-weighted Class Activation Mapping) \cite{selvaraju2017grad} to analyze the auxiliary dataset images that significantly activate the target group outputs. Grad-CAM utilizes the gradients of one or more selected outputs of the model, backpropagating them to the convolutional layers to generate a map of the most activated regions. This process enables the identification of the most significant portions of the image that prompted the model to produce specific outputs. Initially, images from the auxiliary dataset are fed into the model to generate predictions. 
Subsequently, only the outputs corresponding to the target group are filtered. The classes and images generating the highest values are selected for the Grad-CAM application, as they are most likely to confuse the model. Lastly, the images produced by Grad-CAM are included with a heatmap that identifies the activated areas.

\section{Experiments and Results}
\label{sec:experiments}

%Nesta seção, detalhamos a implementação dos modelos e treinamento, bem como as bases de dados utilizadas nos experimentos. Também apresentamos e avaliamos o desempenho de cada modelo em termos de acurácia e curvas ROC/AUC, com variações do hiperparâmetro $\lambda$. Por fim, exibimos as saídas das técnicas de interpretabilidade Layer Correlation e Grad-CAM.

%In this section, we detail the implementation of the models and training, as well as the datasets used in the experiments. We also present and evaluate the performance of each model in terms of accuracy and ROC/AUC curves, with variations of the hyperparameter $\lambda$. Finally, we display the outputs of the interpretability techniques Layer Correlation and Grad-CAM.

%This section details the implementation of the models and training, along with the datasets used in the experiments. The performance of each model in terms of accuracy and ROC/AUC curves is also presented and evaluated, considering variations of the hyperparameter $\lambda$. Lastly, the outputs of the interpretability techniques, Layer Correlation, and Grad-CAM are displayed.

This section describes the experimental design to address our research question. The implementation of the models and training procedures and the datasets employed in the experiments are also detailed. The performance of each model, in terms of accuracy and ROC/AUC curves, is presented and evaluated, considering variations of the hyperparameter $\lambda$. Finally, the results of the interpretability techniques, Layer Correlation, and Grad-CAM are showcased.

In this work, models were implemented using the TensorFlow library and trained on an NVIDIA GeForce GTX TITAN X graphics card with 12GB of memory. The experiments utilized the feature extraction blocks of the InceptionV3 \cite{szegedy2016rethinking} and ResNet50 \cite{he2016deep} models, specifically their convolutional parts. The original dense layers were removed, and two new ones were added. The first dense layer employed 1024 neurons, while the second layer utilized 512 neurons directly connected to the classification layer. AdaGrad \cite{duchi2011adaptive} was chosen as the optimizer, with a learning rate of $1\times10^{-3}$.

All models were pre-trained using the ImageNet dataset, and the weights of the dense layers were initialized with the Glorot Uniform technique \cite{glorot2010understanding}. The base model with InceptionV3 contained $24,397,484$ trainable parameters, whereas ResNet50 had $26,163,724$. The multi-group model increased by $513,000$ parameters, totaling $24,910,484$ for InceptionV3 and $26,676,724$ for ResNet50. When applicable, dropout was applied to the dense layers. In all experiments, the batch size was set to 32 images. For the multi-group model, half of the batch was filled with images from the target dataset and the other half with the auxiliary dataset. The number of training epochs was established at $500$, and the 
hyperparameters vector is fixed in $\mathbf{h} = [\alpha, \beta, \lambda] = [1,1,\lambda]$. In Section~\ref{sec:hyperparameter}, the influence of the hyperparameters $\lambda$ is investigated on simulations.

\subsection{Dataset}

%Para testar a robustez do método proposto em relação à base de dados auxiliar, selecionamos duas bases de dados auxiliares: uma relacionada ao conjunto de dados alvo e outra sem relação. O UFOP Hop Varieties Dataset (UFOP-HVD) \cite{castro2021dataset} foi escolhido como conjunto de dados alvo e consiste em uma base de dados de folhas de lúpulo, uma planta utilizada na fabricação de cerveja. Para o dataset auxiliar relacionado ao domínio do dataset alvo, optamos pelo PlantNet-300K \cite{garcin2021plantnet}, que contém imagens de plantas. O dataset auxiliar não relacionado ao dataset alvo utilizado foi o conhecido ImageNet \cite{deng2009imagenet,russakovsky2015imagenet}, que contém imagens diversas como veículos e animais e é amplamente utilizado em aprendizado de máquina.

%To test the robustness of the proposed method concerning the auxiliary dataset, we selected two auxiliary datasets: one related to the target dataset and another unrelated. The UFOP Hop Varieties Dataset (UFOP-HVD) \cite{castro2021dataset} was chosen as the target dataset and consists of a database of hop leaves, a plant used in beer brewing. For the auxiliary dataset related to the target dataset's domain, we opted for PlantNet-300K \cite{garcin2021plantnet}, which contains images of plants. The unrelated auxiliary dataset used was the well-known ImageNet \cite{deng2009imagenet,russakovsky2015imagenet}, which contains various images such as vehicles and animals and is widely used in machine learning.

To assess the robustness of the proposed method in relation to the auxiliary task, two auxiliary datasets were selected: one related to the target dataset and another unrelated. The UFOP Hop Varieties Dataset (UFOP-HVD) \cite{castro2021dataset} was chosen as the target dataset, consisting of a database of hop leaves, a plant used in beer brewing. PlantNet-300K was selected for the related auxiliary dataset containing images of plants. In contrast, the unrelated auxiliary dataset employed was the well-known ImageNet, which includes various images, such as vehicles and animals, and is widely used in machine learning.

%O UFOP-HVD dataset contém 1592 imagens de folhas de lúpulo jovens e adultas distribuídas em 12 classes/variedades não balanceadas. As fotografias não possuem controle de iluminação, foco, distância e ângulo, e as resoluções variam de 1040x520 a 4096x3072. Para este trabalho, utilizamos a divisão original do dataset, reservando 70\% das imagens para treinamento, 15\% para validação e 15\% para teste. No entanto, em alguns experimentos, reduzimos o conjunto de treinamento para apenas 10\% do seu tamanho original. Algumas amostras do conjunto de dados podem ser visualizadas na Figure \ref{fig:varieties}.

%The UFOP-HVD dataset contains 1592 images of young and mature hop leaves distributed across 12 unbalanced classes/varieties. The photographs have no control over lighting, focus, distance, and angle, and resolutions range from 1040x520 to 4096x3072. In this work, we used the original dataset split, reserving 70\% of the images for training, 15\% for validation, and 15\% for testing. However, in some experiments, we reduced the training set to only 10\% of its original size. Some samples from the dataset can be viewed in Figure \ref{fig:varieties}.

\begin{figure}[!ht]
\centering
%\hspace*{\fill}
\subfigure[]{\includegraphics[width = 0.7in,height=0.94in]{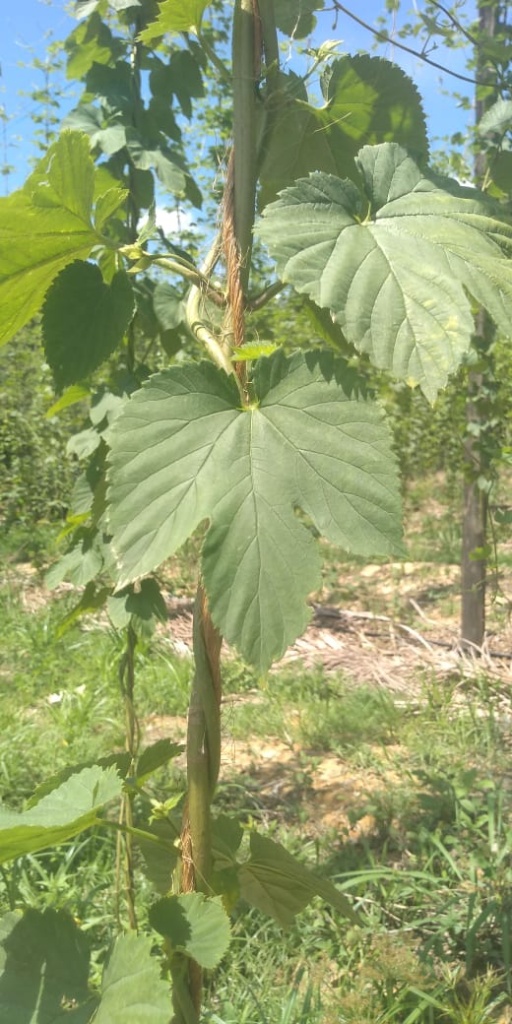}\label{fig:cascade}} 
%\hfill
\subfigure[]{\includegraphics[angle=90,width = 0.7in,height=0.94in]{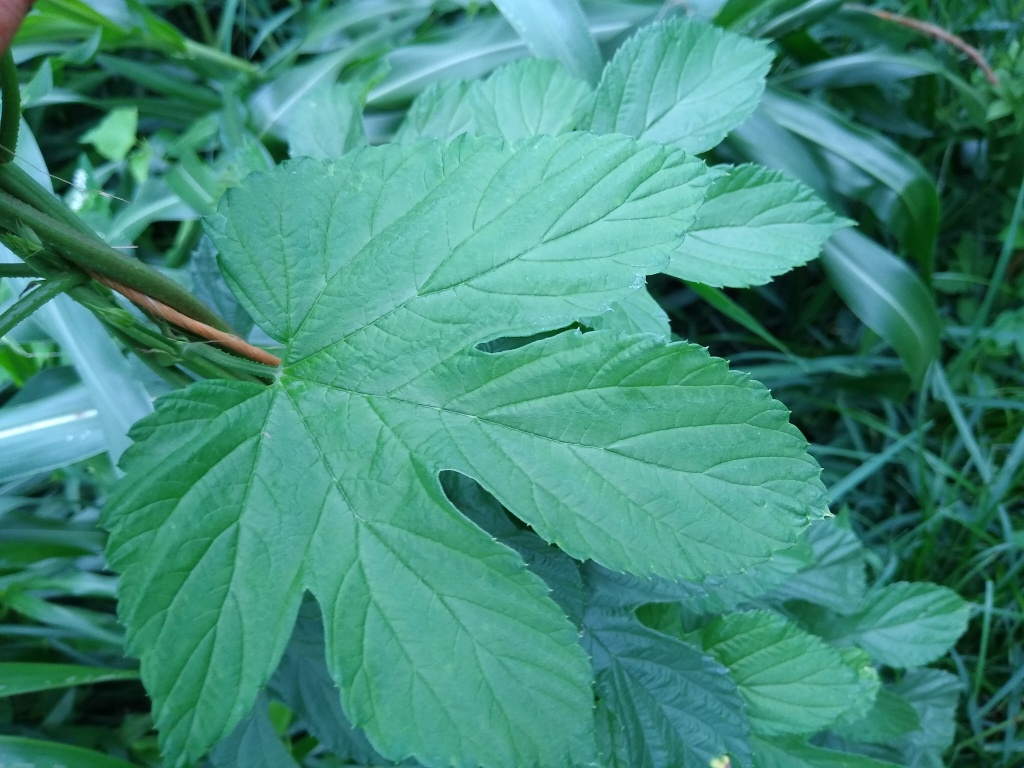}\label{fig:centennial}}
%\hfill
\subfigure[]{\includegraphics[width = 0.7in,height=0.94in]{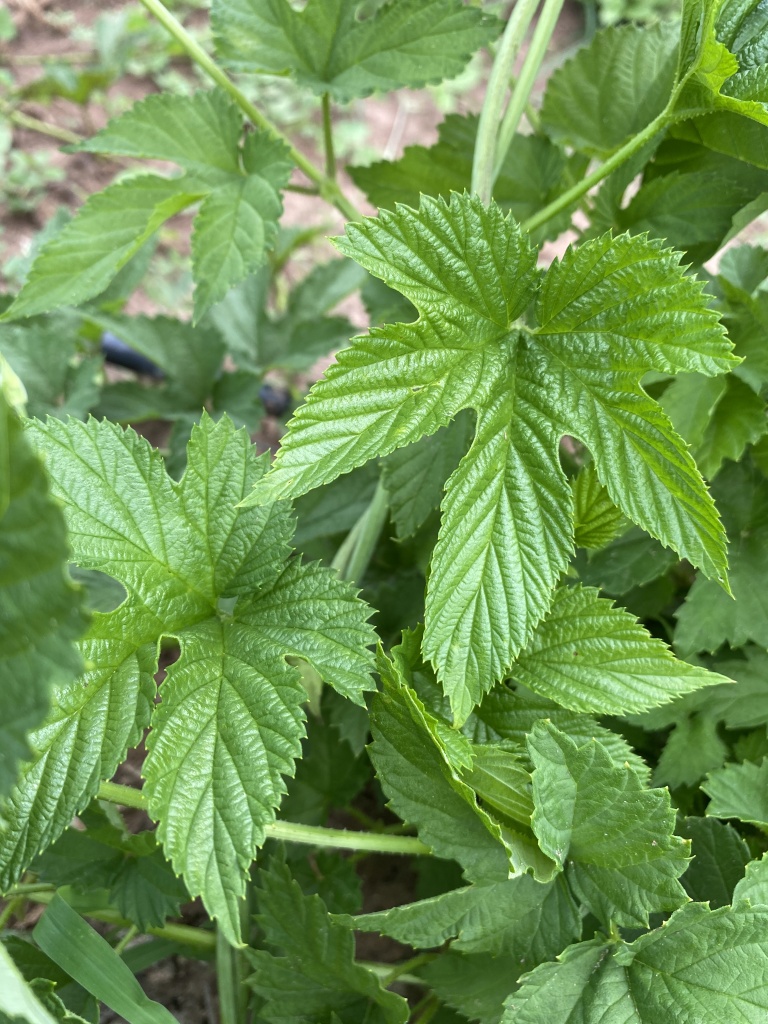}\label{fig:cluster}}
%\hfill
\subfigure[]{\includegraphics[width = 0.7in,height=0.94in]{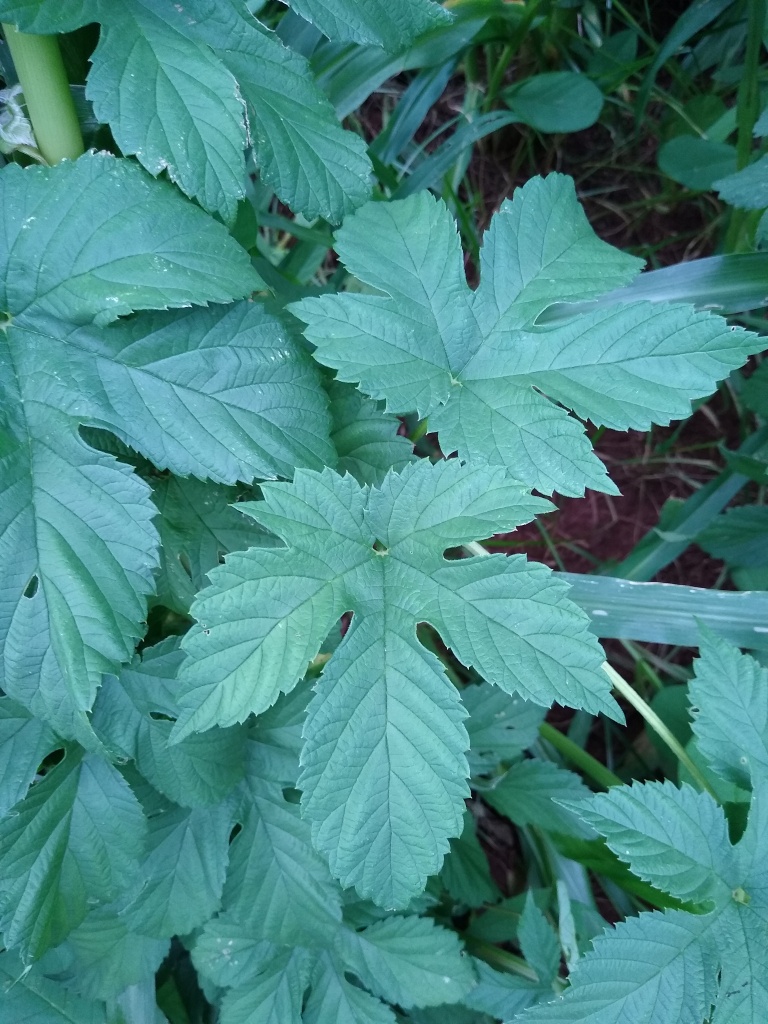}\label{fig:comet}}
%\hspace*{\fill}
%\hspace*{\fill}
\subfigure[]{\includegraphics[angle=90,width = 0.7in,height=0.94in]{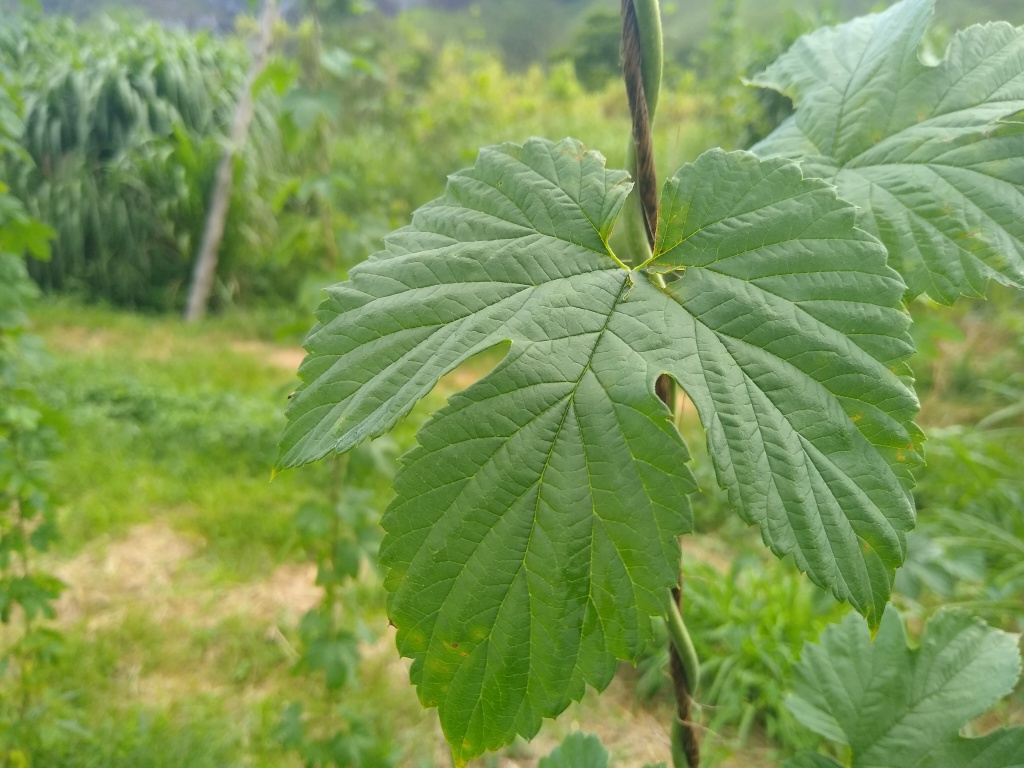}\label{fig:hallertau_mittelfrueh}} 
%\hfill
\subfigure[]{\includegraphics[width = 0.7in,height=0.94in]{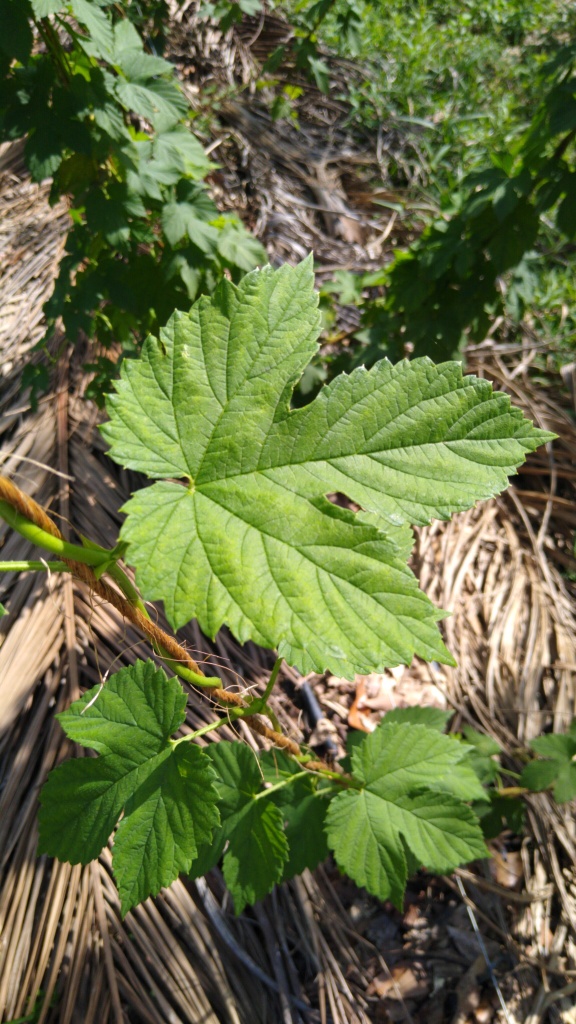}\label{fig:nugget}}
%\hfill
\\
\subfigure[]{\includegraphics[width = 0.7in,height=0.94in]{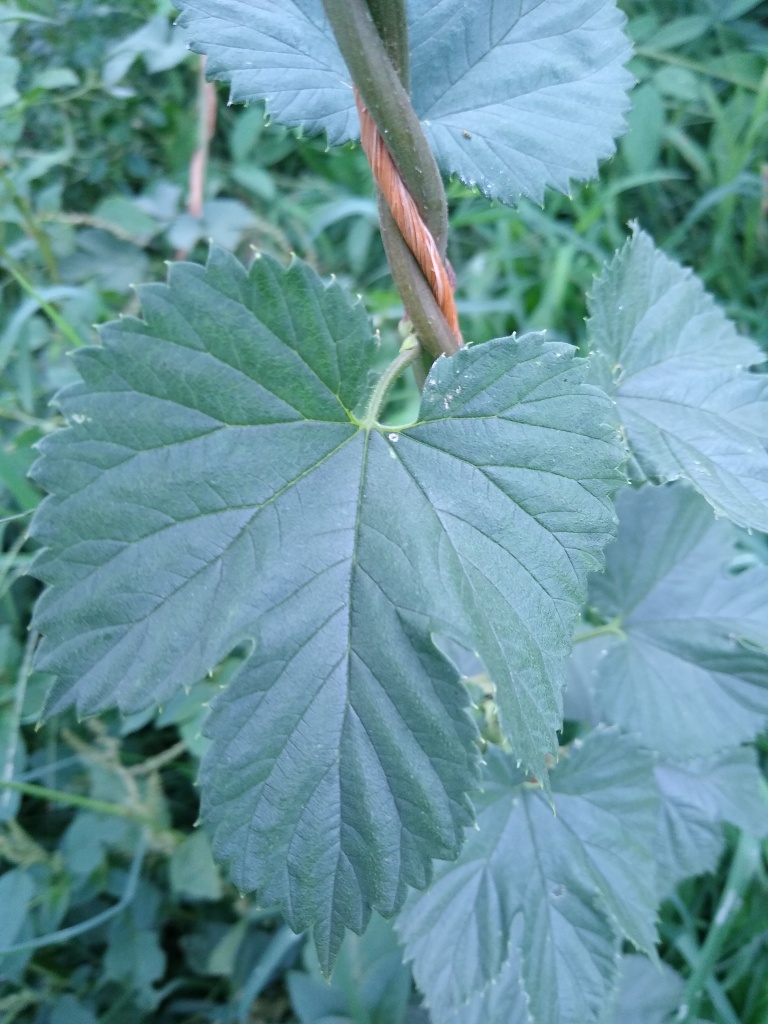}\label{fig:saaz}}
%\hfill
\subfigure[]{\includegraphics[width = 0.7in,height=0.94in]{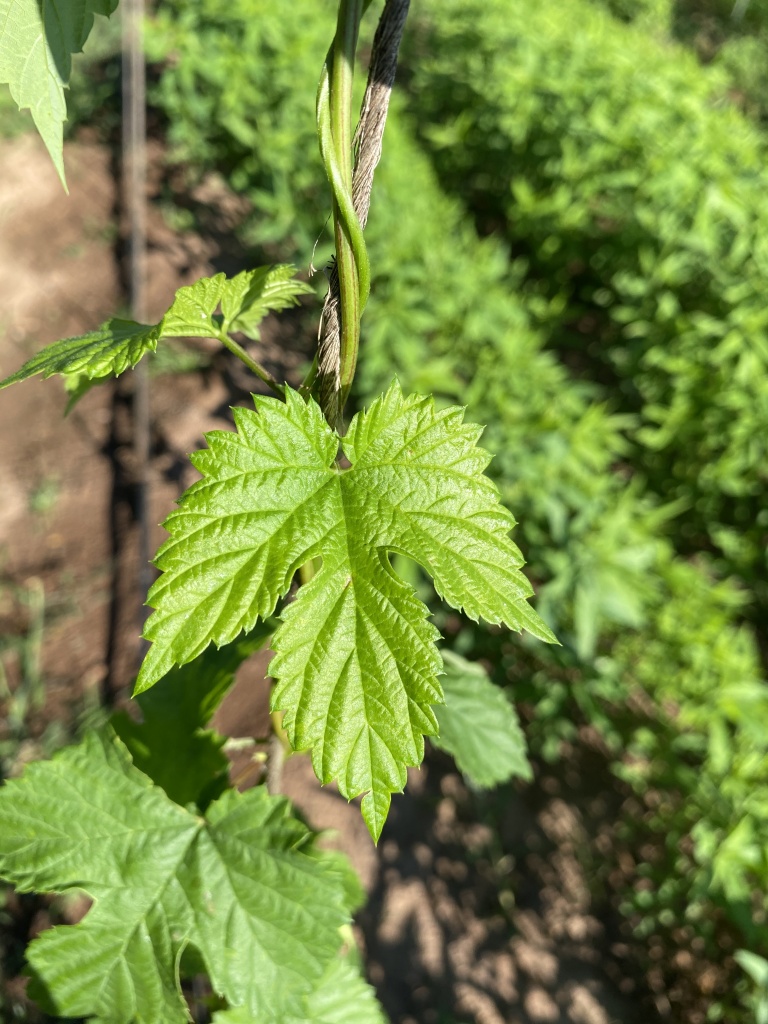}\label{fig:sorachi_ace}}
%\hspace*{\fill}
%\hspace*{\fill}
\subfigure[]{\includegraphics[angle=90,width = 0.7in,height=0.94in]{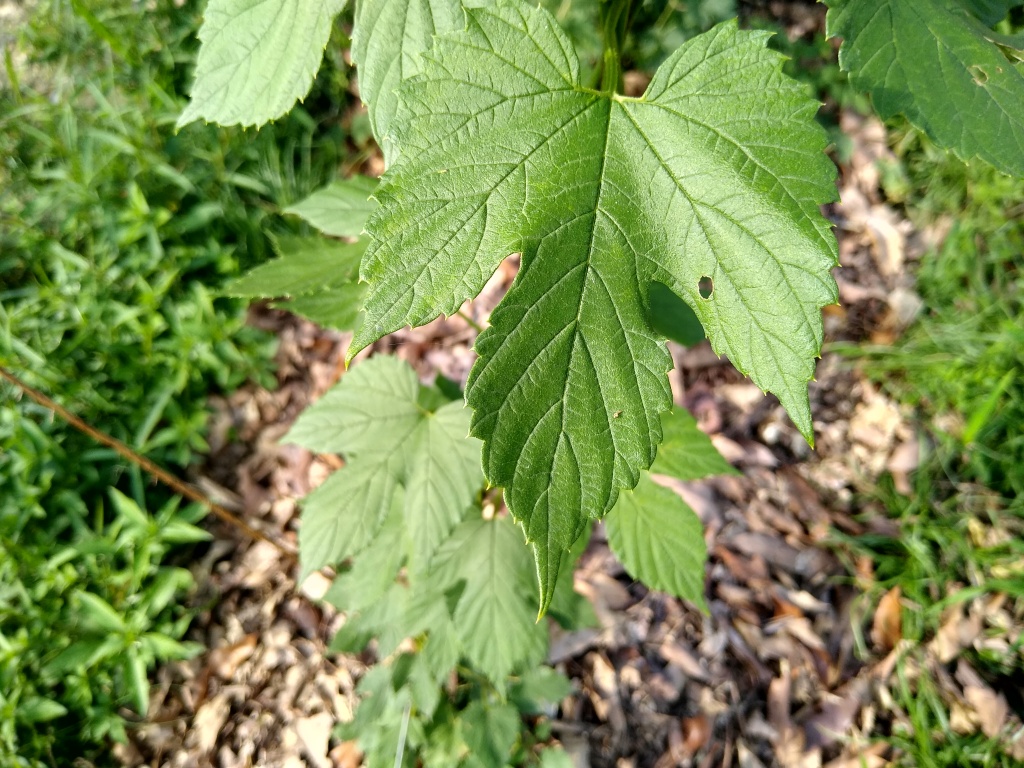}\label{fig:tahoma}} 
%\hfill
\subfigure[]{\includegraphics[width = 0.7in,height=0.94in]{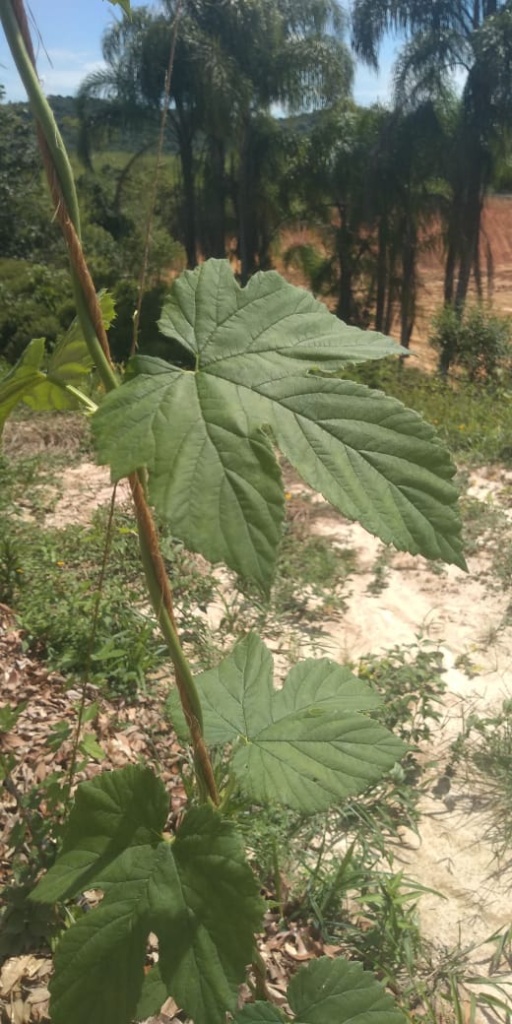}\label{fig:triple_pearl}}
%\hfill
\subfigure[]{\includegraphics[width = 0.7in,height=0.94in]{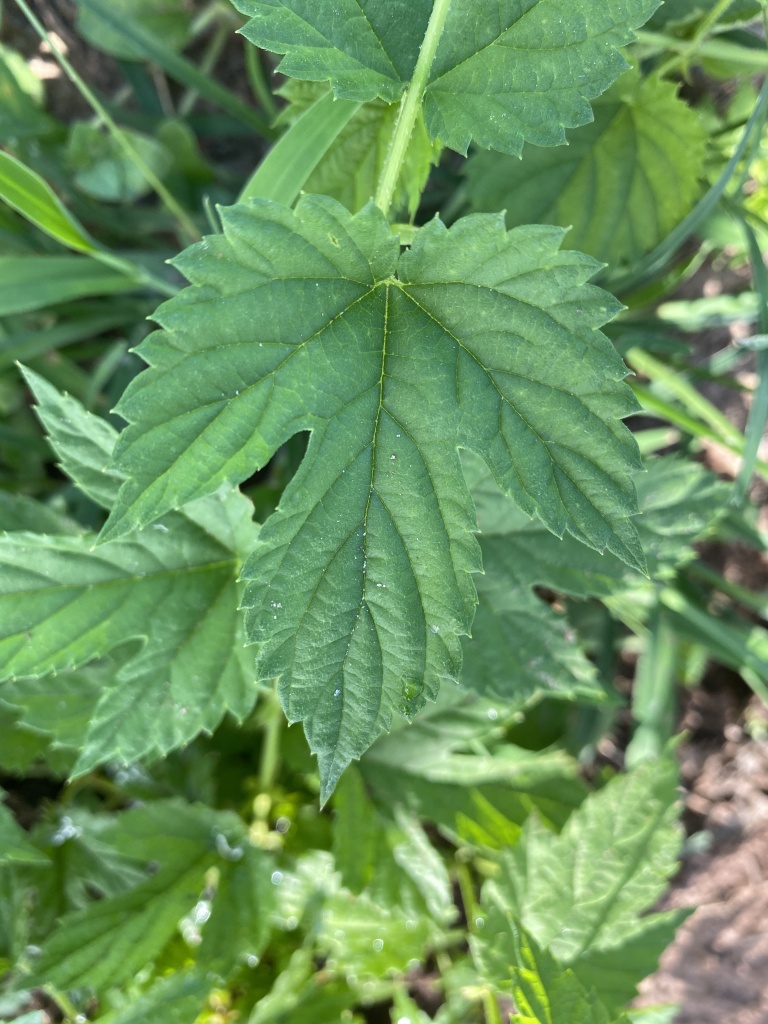}\label{fig:triumph}}
%\hfill
\subfigure[]{\includegraphics[angle=90,width = 0.7in,height=0.94in]{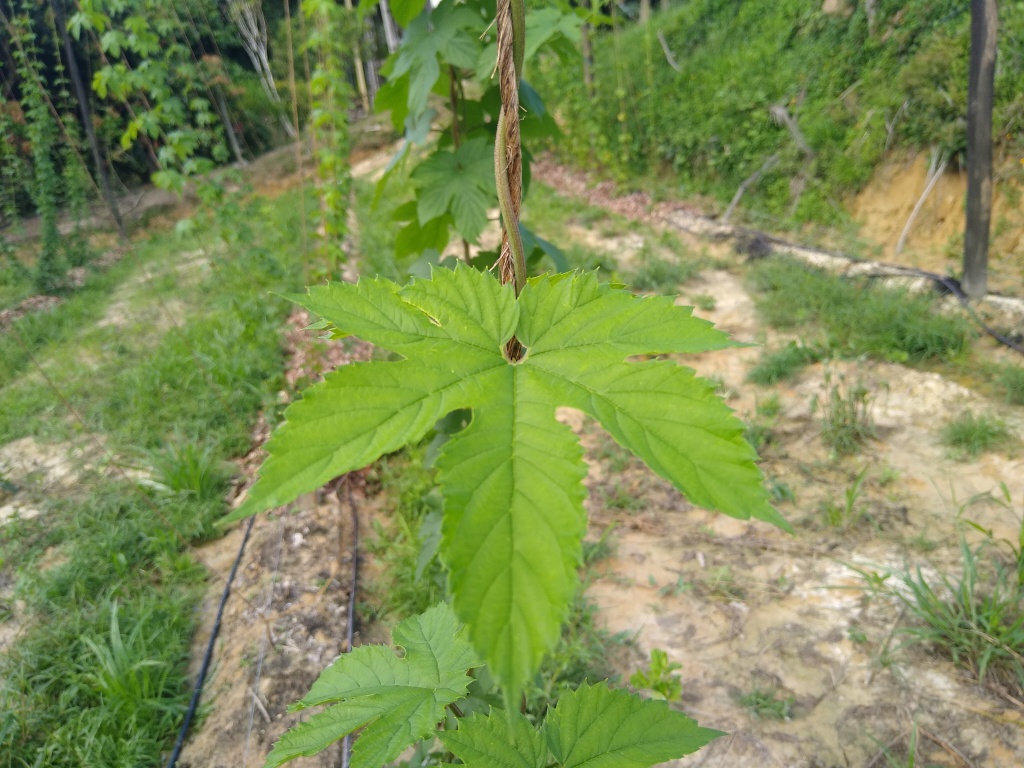}\label{fig:zeus}}
%\hspace*{\fill}
\caption{Examples of the 12 hop varieties used in this work: (a) Cascade; (b) Centennial; (c) Cluster; (d) Comet; (e) Hallertau Mittelfrueh; (f) Nugget; (g) Saaz; (h) Sorachi Ace; (i) Tahoma; (j) Triple Pearl; (k) Triumph; (l) Zeus.}
\label{fig:varieties}
\end{figure}

The UFOP-HVD dataset comprises 1592 images of young and mature hop leaves distributed across 12 unbalanced classes/varieties. The photographs exhibit no control over lighting, focus, distance, and angle, with resolutions ranging from $1040\times520$ to $4096\times3072$. In this work, the original dataset split was utilized, allocating 70\% of the images for training, 15\% for validation, and 15\% for testing. However, in some experiments, the training set was reduced to only 10\% of its original size. Several samples from the dataset can be observed in Figure \ref{fig:varieties}.

%O dataset PlantNet-300K é composto por $1,081$ espécies de plantas e $306,146$ imagens, sendo altamente desbalanceado devido ao fato de algumas espécies possuírem poucas amostras. Selecionamos as $1,000$ classes com menos exemplos, totalizando $72,680$ imagens para o conjunto de treinamento. Algumas amostras do conjunto de dados podem ser conferidas na Figure \ref{fig:plantnet}.

%The PlantNet-300K dataset comprises $1,081$ plant species and $306,146$ images, being highly unbalanced due to some species having few samples. We selected the $1,000$ classes with the fewest examples, totaling $72,680$ images for the training set. Some samples from the dataset can be seen in Figure \ref{fig:plantnet}.

\begin{figure}[!ht]
\centering
\includegraphics[width = .46\linewidth]{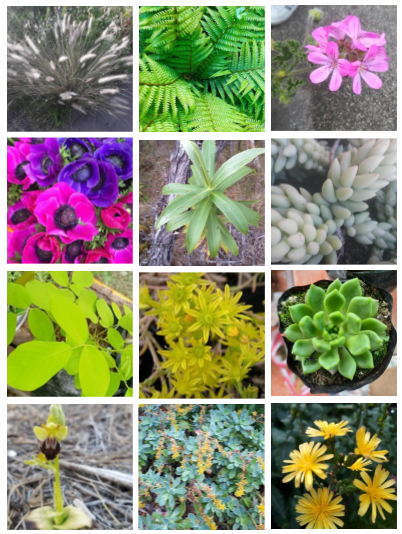} 
\caption{Examples of plant species from the PlantNet-300K dataset.}
\label{fig:plantnet}
\end{figure}

Composed of $1,081$ plant species and $306,146$ images, the PlantNet-300K dataset is highly unbalanced, as some species have limited samples. The $1,000$ classes with the fewest examples were selected, resulting in $72,680$ images for the training set. A selection of samples from the dataset can be viewed in Figure \ref{fig:plantnet}.

%O ImageNet é um dataset com mais de $14,197,122$ imagens, já divididas em conjunto de treinamento, validação e teste, distribuídas em 1000 classes. Possui temas variados, como animais, veículos, pessoas, roupas, alimentos, jogos e equipamentos de tecnologia, entre outros. Neste trabalho, utilizamos um subconjunto com 100 amostras de cada classe, obtidas aleatoriamente, totalizando $100,000$ imagens. A Figure \ref{fig:imagenet} contém algumas amostras do dataset.

%ImageNet is a dataset with over $14,197,122$ images, already split into training, validation, and testing sets, distributed across 1000 classes. It features a wide variety of themes, such as animals, vehicles, people, clothing, food, games, and technology equipment, among others. In this work, we used a subset with 100 samples from each class, obtained randomly, totaling $100,000$ images. Figure \ref{fig:imagenet} contains some samples from the dataset.

\begin{figure}[!ht]
\centering
\includegraphics[width = .46\linewidth]{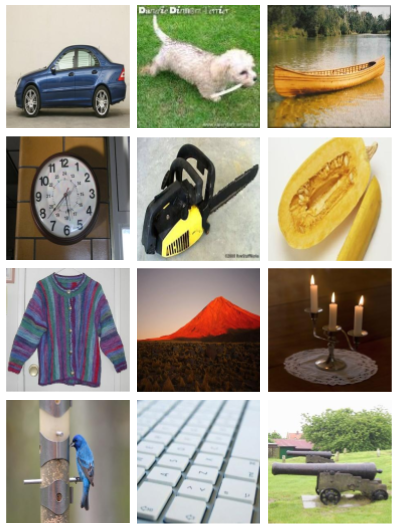} 
\caption{Examples of ImageNet images, from left to right and top to bottom: car, dog, boat, clock, chainsaw, pumpkin, sweater, volcano, candle, bird, keyboard, and cannon.}
\label{fig:imagenet}
\end{figure}

ImageNet, a dataset of over $14,197,122$ images, is distributed across 1,000 classes and divided into training, validation, and testing sets. The dataset encompasses a diverse range of themes, including animals, vehicles, people, clothing, food, games, and technology equipment, among others. In this study, a subset containing 100 random samples from each class was utilized, amounting to $100,000$ images. A selection of samples from the dataset can be viewed in Figure \ref{fig:imagenet}.

\subsection{Sensitivity analysis of hyperparameter}
\label{sec:hyperparameter}

%O primeiro experimento teve como objetivo estudar o comportamento do hiperparâmetro $\lambda$ para cada modelo e dataset auxiliar. Variamos $\lambda$ de $0.1$ a $1.0$ em intervalos de $0.1$. A Figure \ref{fig:lambda_variation} apresenta os gráficos de acurácia dos modelos na base de validação em função da variação de $\lambda$. A linha tracejada em vermelho representa o resultado do modelo base, quando nenhuma regularização é aplicada, enquanto os pontos em azul e verde se referem à técnica Simultaneous Learning utilizando o ImageNet e o PlantNet-300K como base auxiliar, respectivamente. Todas as combinações de modelo, dataset auxiliar e $\lambda$ promoveram algum aumento na acurácia. É importante observar que houve um ganho expressivo no desempenho da InceptionV3, que se aproximou do desempenho da ResNet50, o que não ocorre no modelo base. Também foi observada uma tendência de melhoria do desempenho à medida que $\lambda$ aumenta. Os melhores valores de $\lambda$ encontrados foram $0.7$ para ResNet50 com ImageNet, $1.0$ para ResNet50 com PlantNet-300K, $0.7$ para InceptionV3 com ImageNet e $0.9$ para InceptionV3 com PlantNet-300K. Cabe ressaltar que quando $\lambda$ é igual a $1$, o modelo ignora a classificação do dataset auxiliar, mantendo apenas a penalização para erros inter-grupos. Parece que os modelos com dataset auxiliar mais relacionado ao dataset alvo se beneficiam menos desta classificação.

\begin{figure}[!ht]
%\centering
%\hspace*{\fill}
\subfigure[Model: ResNet50]{
\includegraphics[width = .46\linewidth]{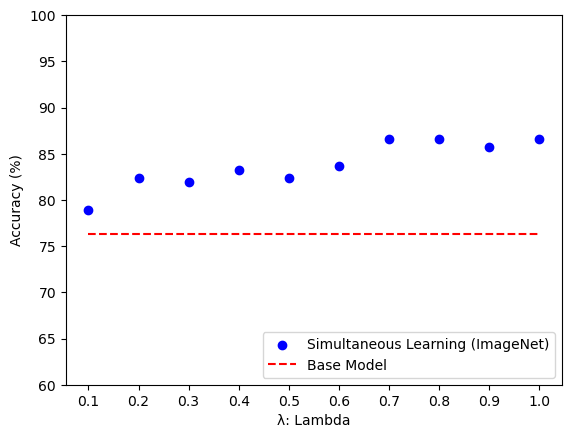}
\label{fig:lambda_variation_resnet50_imagenet}} 
%\\
\subfigure[Model: ResNet50]{
\includegraphics[width = .46\linewidth]{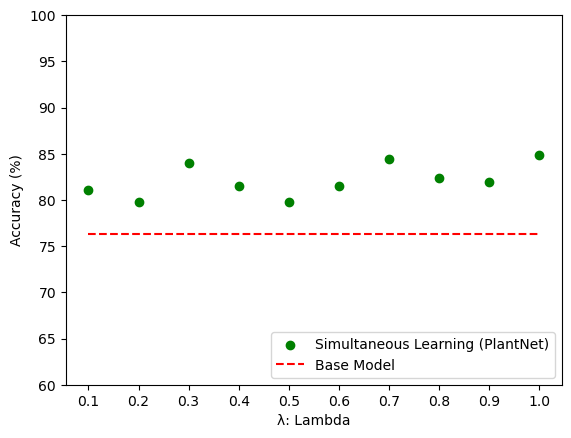}
\label{fig:lambda_variation_resnet50_plantnet}}
\\
\subfigure[Model: InceptionV3]{
\includegraphics[width = .46\linewidth]{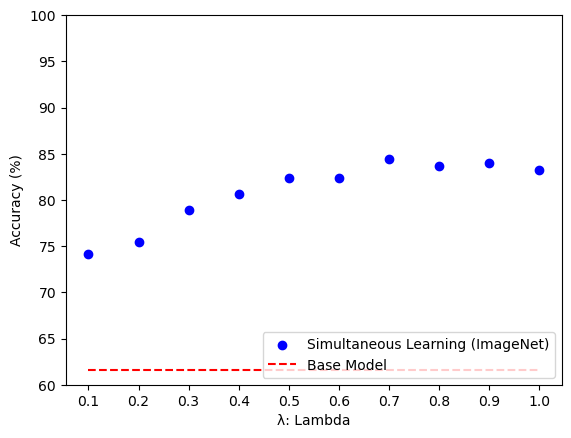}
\label{fig:lambda_variation_inceptionv3_imagenet}} 
%\\
\subfigure[Model: InceptionV3]{
\includegraphics[width = .46\linewidth]{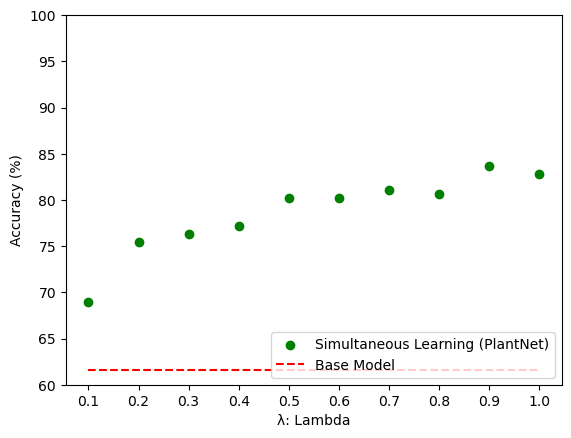}
\label{fig:lambda_variation_inceptionv3_plantnet}}
%\hspace*{\fill}
\caption{Delimited accuracy (vertical axis) on the validation set as a function of $\lambda$ (horizontal axis). The dashed red line represents the result of the base model when no regularization is applied, and the blue and green points indicate the performance of the Simultaneous Learning technique using ImageNet and PlantNet as auxiliary bases, respectively.}
\label{fig:lambda_variation}
\end{figure}

The first experiment aimed to study the behavior of the hyperparameter $\lambda$ for each model and auxiliary dataset. It varied from $0.1$ to $1.0$ in increments of $0.1$. Figure \ref{fig:lambda_variation} presents the accuracy plots of the models on the validation set as a function of $\lambda$ variation. The red dashed line represents the result of the base model, where no regularization is applied, while the blue and green points refer to the Simultaneous Learning technique using ImageNet and PlantNet-300K as auxiliary datasets, respectively. All combinations of the model, auxiliary dataset, and $\lambda$ promoted some increase in accuracy. It is important to note that there was a significant performance gain for InceptionV3, which approached the performance of ResNet50, which does not occur in the base model. An improvement trend was also observed as $\lambda$ increases. The best $\lambda$ values found were $0.7$ for ResNet50 with ImageNet, $1.0$ for ResNet50 with PlantNet-300K, $0.7$ for InceptionV3 with ImageNet, and $0.9$ for InceptionV3 with PlantNet-300K. It is worth mentioning that when $\lambda$ is equal to $1$, the model ignores the classification of the auxiliary dataset, keeping only the penalty for inter-group errors. It seems that models with an auxiliary dataset more related to the target dataset benefit less from this classification.

\subsection{Performance}

Subsequent experiments were performed on the test dataset, with the hyperparameter $\lambda$ set to the optimal values identified in the previous investigation. Tables \ref{tab:acc_resnet50} and \ref{tab:acc_inceptionv3} display the accuracy of the ResNet50 and InceptionV3 models, respectively. The performance of the base models without regularization, which served as a baseline, is presented alongside the performance of the base models with dropout applied, using dropout rates of $0.2$, $0.5$, and $0.8$, indicated in parentheses. The Simultaneous Learning technique's performance, utilizing both auxiliary datasets, is significantly superior to the baseline and dropout, exhibiting at least a 5-percentage point difference in all comparisons and attaining up to 22 points higher than the baseline with InceptionV3.

\begin{table}[!ht]
\centering
\caption{Comparison of ResNet50 model performance without regularization, with dropout, and with Simultaneous Learning technique.}
\begin{tabular}{lc}
\toprule
Model                                       & Accuracy (\%) \\
\midrule
Baseline                    & 76.29 \\
Dropout (0.2)               & 77.16 \\
Dropout (0.5)               & 78.02 \\
Dropout (0.8)               & 81.03 \\
\textbf{SL (PlantNet, $\lambda$ = 1.0)} & \textbf{86.21} \\
\textbf{SL (ImageNet, $\lambda$ = 0.7)} & \textbf{87.93} \\
\bottomrule
\end{tabular}
\label{tab:acc_resnet50}
\end{table}

\begin{table}[!ht]
\centering
\caption{Comparison of InceptionV3 model performance without regularization, with dropout, and with Simultaneous Learning technique.}
\begin{tabular}{lc}
\toprule
Model                                       & Accuracy (\%) \\
\midrule
Baseline                    & 61.21 \\
Dropout (0.2)               & 57.76 \\
Dropout (0.5)               & 67.24 \\
Dropout (0.8)               & 66.38 \\
\textbf{SL (PlantNet, $\lambda$ = 0.9)} & \textbf{82.76} \\
\textbf{SL (ImageNet, $\lambda$ = 0.7)} & \textbf{83.62} \\
\bottomrule
\end{tabular}
\label{tab:acc_inceptionv3}
\end{table}

%Verificamos o desempenho do método Simultaneous Learning em bases de dados muito pequenas, reduzindo a base de treinamento para apenas $10\%$ de seu tamanho original, retreinando os modelos e aplicando as mesmas regularizações. Os resultados estão nas tabelas \ref{tab:acc_resnet50_reduced} e \ref{tab:acc_inceptionv3_reduced}. Novamente, o Simultaneous Learning apresentou desempenho superior. Em alguns casos, o dropout chegou a piorar a acurácia em relação ao baseline.

%We assessed the performance of the Simultaneous Learning method on very small datasets by sub-sampling the training set to just $10\%$ of its original size, retraining the models, and applying the same regularizations. The results are shown in tables \ref{tab:acc_resnet50_reduced} and \ref{tab:acc_inceptionv3_reduced}. Once again, Simultaneous Learning demonstrated superior performance. In some cases, dropout even resulted in worse accuracy compared to the baseline.

The performance of the Simultaneous Learning method was evaluated on considerably smaller datasets by sub-sampling the training set to merely $10\%$ of its original size, retraining the models, and applying the same regularizations. The results are in Tables \ref{tab:acc_resnet50_reduced} and \ref{tab:acc_inceptionv3_reduced}. Simultaneous Learning once again exhibited superior performance. However, in certain instances, the application of dropout led to a decline in accuracy compared to the baseline.

\begin{table}[!ht]
\centering
\caption{Comparison of ResNet50 model performance on a reduced dataset without regularization, with dropout, and with Simultaneous Learning technique.}
\begin{tabular}{lc}
\toprule
Model                                       & Accuracy (\%) \\
\midrule
Baseline                    & 45.69 \\
Dropout (0.2)               & 46.12 \\
Dropout (0.5)               & 43.97 \\
Dropout (0.8)               & 43.97 \\
\textbf{SL (PlantNet, $\lambda$ = 1.0)} & \textbf{51.72} \\
\textbf{SL (ImageNet, $\lambda$ = 0.7)} & \textbf{56.03} \\
\bottomrule
\end{tabular}
\label{tab:acc_resnet50_reduced}
\end{table}

\begin{table}[!ht]
\centering
\caption{Comparison of InceptionV3 model performance on a reduced dataset without regularization, with dropout, and with Simultaneous Learning technique.}
\begin{tabular}{lc}
\toprule
Model                                       & Accuracy (\%) \\
\midrule
Baseline                    & 27.16 \\
Dropout (0.2)               & 18.10 \\
Dropout (0.5)               & 29.31 \\
Dropout (0.8)               & 28.45 \\
\textbf{SL (PlantNet, $\lambda$ = 0.9)} & \textbf{41.38} \\
\textbf{SL (ImageNet, $\lambda$ = 0.7)} & \textbf{42.67} \\
\bottomrule
\end{tabular}
\label{tab:acc_inceptionv3_reduced}
\end{table}

\subsection{Combined Performance}

%É de interesse do nosso estudo investigar a viabilidade de combinar mais de uma técnica de regularização. Neste sentido, combinamos o método Simultaneous Learning com o dropout e os resultados podem ser vistos nas tabelas \ref{tab:acc_resnet50_combined} e \ref{tab:acc_inceptionv3_combined}. Para a ResNet50, obtivemos um resultado superior às técnicas isoladas quando o dropout é de $0.2$ para ambos datasets auxiliares. Já para a InceptionV3, a melhoria na acurácia ocorreu apenas no dataset ImageNet com o dropout de $0.8$. Com o PlantNet e dropout de $0.8$, o resultado se manteve constante.

%It is interesting for our study to investigate the feasibility of combining more than one regularization technique. In this regard, we combined the Simultaneous Learning method with dropout, and the results can be seen in tables \ref{tab:acc_resnet50_combined} and \ref{tab:acc_inceptionv3_combined}. For ResNet50, we obtained a higher result than the isolated techniques when the dropout rate was $0.2$ for both auxiliary datasets. For InceptionV3, the improvement in accuracy occurred only with the ImageNet dataset and a dropout rate of $0.8$. The result remained constant with the PlantNet and a dropout rate of $0.8$.

In the context of our study, examining the feasibility of combining multiple regularization techniques is of interest. To this end, the Simultaneous Learning method was integrated with dropout, and the results are displayed in Tables \ref{tab:acc_resnet50_combined} and \ref{tab:acc_inceptionv3_combined}. For the ResNet50, a superior outcome was achieved compared to the individual techniques when the dropout rate was set at $0.2$ for both auxiliary datasets. For InceptionV3, the improvement in accuracy transpired only with the ImageNet dataset and a dropout rate of $0.8$. The result remained unchanged with the PlantNet dataset and a dropout rate of $0.8$.

\begin{table}[!ht]
\centering
\caption{Comparison of ResNet50 model performance, combining Simultaneous Learning with dropout.}
\begin{tabular}{lc}
\toprule
Model                                       & Accuracy (\%) \\
\midrule
\textbf{SL (PlantNet, $\lambda$ = 1.0) + Dropout (0.2)} & \textbf{89.66} \\
SL (PlantNet, $\lambda$ = 1.0) + Dropout (0.5) & 84.05 \\
SL (PlantNet, $\lambda$ = 1.0) + Dropout (0.8) & 81.03 \\
\textbf{SL (ImageNet, $\lambda$ = 0.7) + Dropout (0.2)} & \textbf{88.36} \\
SL (ImageNet, $\lambda$ = 0.7) + Dropout (0.5) & 84.05 \\
SL (ImageNet, $\lambda$ = 0.7) + Dropout (0.8) & 80.17 \\
\bottomrule
\end{tabular}
\label{tab:acc_resnet50_combined}
\end{table}

\begin{table}[!ht]
\centering
\caption{Comparison of InceptionV3 model performance, combining Simultaneous Learning with dropout.}
\begin{tabular}{lc}
\toprule
Model                                       & Accuracy (\%) \\
\midrule
SL (PlantNet, $\lambda$ = 0.9) + Dropout (0.2) & 83.19 \\
SL (PlantNet, $\lambda$ = 0.9) + Dropout (0.5) & 81.47 \\
\textbf{SL (PlantNet, $\lambda$ = 0.9) + Dropout (0.8)} & \textbf{83.62} \\
SL (ImageNet, $\lambda$ = 0.7) + Dropout (0.2) & 84.48 \\
SL (ImageNet, $\lambda$ = 0.7) + Dropout (0.5) & 82.76 \\
\textbf{SL (ImageNet, $\lambda$ = 0.7) + Dropout (0.8)} & \textbf{87.07} \\
\bottomrule
\end{tabular}
\label{tab:acc_inceptionv3_combined}
\end{table}

\subsection{ROC Curves / AUC}

%Considerando que a base de dados alvo não é balanceada, é importante avaliar o desempenho do modelo em termos de sensibilidade e especificidade. Para isso, foram construídas as curvas ROC e AUC, conforme ilustrado na Figura \ref{fig:roc_auc}, para os modelos ResNet50 e InceptionV3 aplicados tanto na base de dados completa quanto na reduzida. A linha vermelha representa o baseline, enquanto as linhas azul e verde correspondem, respectivamente, ao Simultaneous Learning com ImageNet e ao Simultaneous Learning com PlantNet. Os resultados mostram que o Simultaneous Learning apresentou desempenho superior ou, no mínimo, igual ao baseline em toda a curva, sendo que a AUC também foi maior em todos os casos.

%Considering that the target dataset is unbalanced, it is important to evaluate the performance of the model in terms of sensitivity and specificity. For this purpose, ROC curves and AUC were constructed, as illustrated in Figure \ref{fig:roc_auc}, for the ResNet50 and InceptionV3 models applied to both the complete and reduced datasets. The red line represents the baseline, while the blue and green lines correspond to Simultaneous Learning with ImageNet and Simultaneous Learning with PlantNet. The results show that Simultaneous Learning exhibited superior or, at the very least, equal performance to the baseline across the entire curve, with the AUC being higher in all cases.

\begin{figure}[!ht]
\centering
%\hspace*{\fill}
\subfigure[Model: ResNet50]{
\includegraphics[width = .46\linewidth]{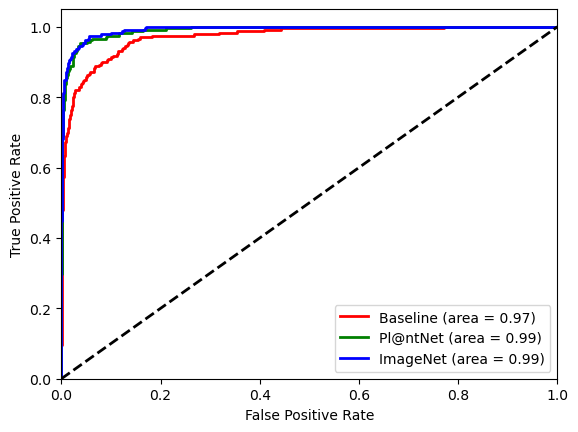}
\label{fig:roc_auc_resnet50}} 
%\\
\subfigure[Model: InceptionV3]{
\includegraphics[width = .46\linewidth]{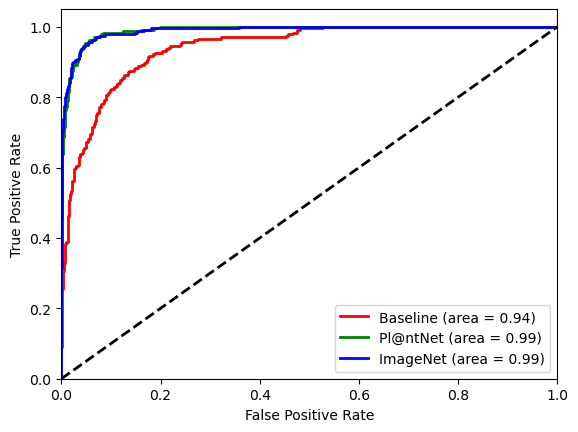}
\label{fig:roc_auc_inception_v3}}
\\
\subfigure[Model: ResNet50 (Reduced dataset)]{
\includegraphics[width = .46\linewidth]{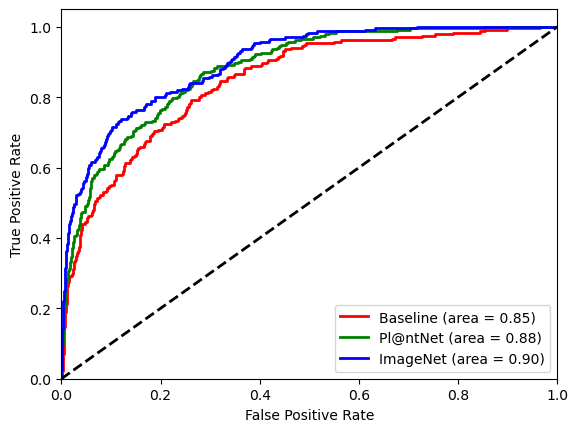}
\label{fig:roc_auc_resnet50_reduced10}} 
%\\
\subfigure[Model: InceptionV3 (Reduced dataset)]{
\includegraphics[width = .46\linewidth]{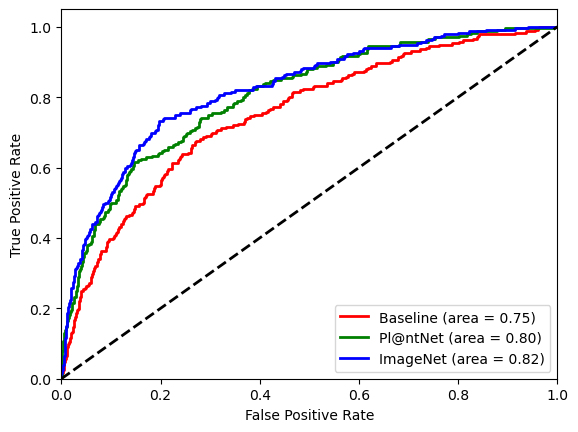}
\label{fig:roc_auc_inception_v3_reduced10}}
%\hspace*{\fill}
\caption{ROC curve and AUC of ResNet50 and InceptionV3 models. The red line corresponds to the baseline model. The green line applies Simultaneous Learning with the PlantNet dataset as the auxiliary base, and the blue line uses Simultaneous Learning with ImageNet.}
\label{fig:roc_auc}
\end{figure}

Given the unbalanced nature of the target dataset, evaluating the model's performance in terms of sensitivity and specificity is crucial. To this end, ROC (Receiver Operating Characteristic) curves and AUC (Area under the ROC Curve) were constructed, as depicted in Figure \ref{fig:roc_auc}, for both the ResNet50 and InceptionV3 models applied to the complete and reduced datasets. The red line represents the baseline, while the blue and green lines correspond to Simultaneous Learning with ImageNet and Simultaneous Learning with PlantNet, respectively. The results demonstrate that Simultaneous Learning displayed superior or, at the very least, equivalent performance to the baseline across the entire curve, with the AUC consistently higher in all cases.

\subsection{Interpretability}

\begin{figure}[!ht]
\centering
\subfigure[5 main layers]{
\includegraphics[width = .46\linewidth]{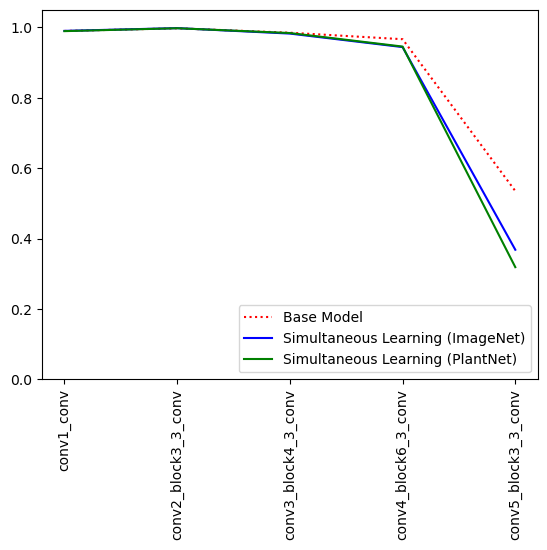}
\label{fig:layer_correlation_layers_blocks_end}} 
%\\
\subfigure[10 last layers]{
\includegraphics[width = .46\linewidth]{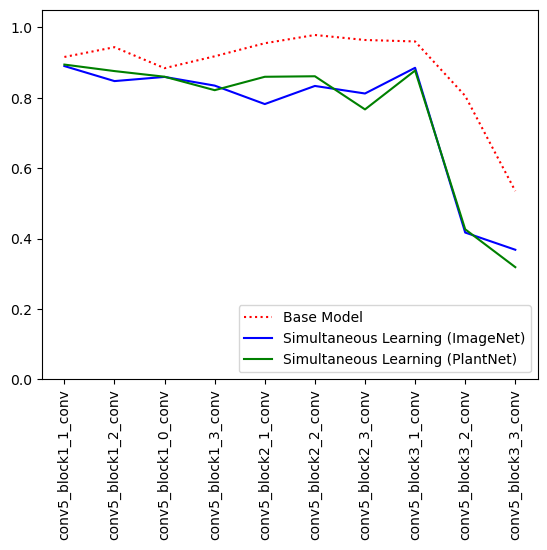}
\label{fig:layer_correlation_layers_layers_last_block}}
\caption{(a) Layer Correlation among class representations in the final layer of each of the five groups within the ResNet50 architecture. (b) Inter-class Layer Correlation within the last 10 layers of the ResNet50 model.}
\label{fig:layer_correlation}
\end{figure}

In this study, the Layer Correlation technique is applied to both the base and multi-group models. The evaluation is conducted solely on the ResNet50 network, which contains 53 convolutional layers divided into five groups. For enhanced visualization, Figure \ref{fig:layer_correlation_layers_blocks_end} displays the correlation of the final layer in each group. A dramatic reduction in the correlation between classes in the last layer can be observed for the Simultaneous Learning method, with both ImageNet and PlantNet as auxiliary datasets. This finding justifies its superior performance, as less correlated features yield more pertinent information for class determination. Figure \ref{fig:layer_correlation_layers_layers_last_block} provides a view of the network's final 10 layers, revealing that Simultaneous Learning generated less correlated features in all of them.

\begin{figure}[!ht]
\centering
\includegraphics[width = .24\linewidth]{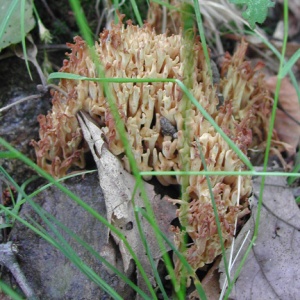}
\hspace{-3mm}
\includegraphics[width = .24\linewidth]{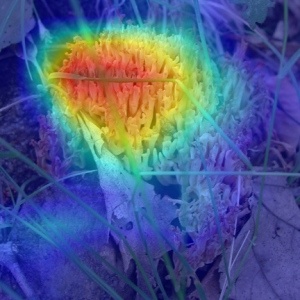}
\hspace{-3mm}
\includegraphics[width = .24\linewidth]{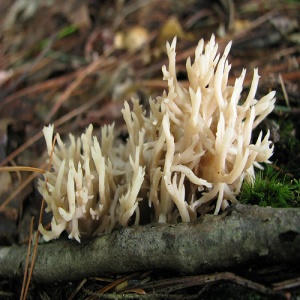}
\hspace{-3mm}
\includegraphics[width = .24\linewidth]{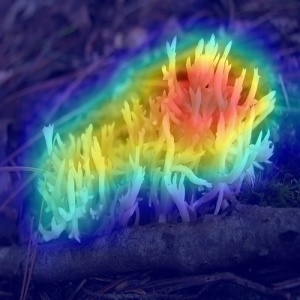}
\\
\vspace{-1mm}
\includegraphics[width = .24\linewidth]{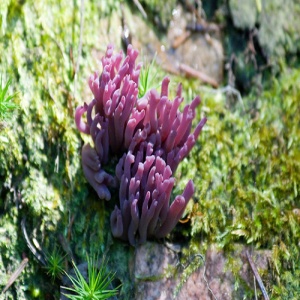}
\hspace{-3mm}
\includegraphics[width = .24\linewidth]{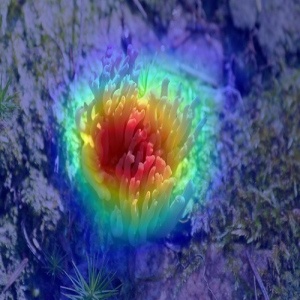}
\hspace{-3mm}
\includegraphics[width = .24\linewidth]{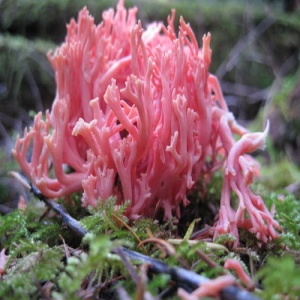}
\hspace{-3mm}
\includegraphics[width = .24\linewidth]{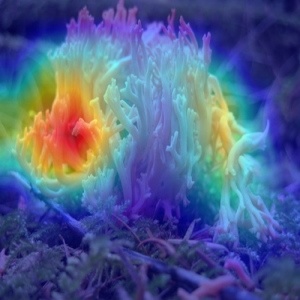}
\caption{Examples of images from the Coral Fungus class drawn from the auxiliary dataset. Corresponding heatmaps are also displayed, which represent the regions that most strongly activated the outputs of the target group. These heatmaps were generated by employing the Grad-CAM method.}
\label{fig:gradcam_coral_fungus}
\end{figure}

\begin{figure}[!ht]
\centering
\includegraphics[width = .24\linewidth]{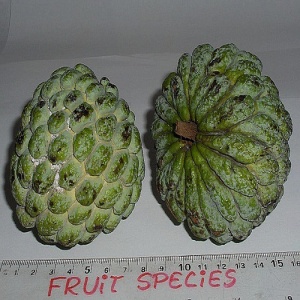}
\hspace{-3mm}
\includegraphics[width = .24\linewidth]{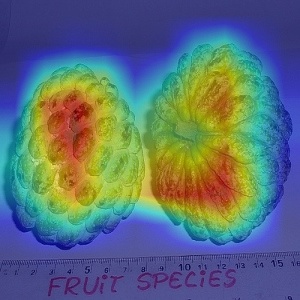}
\hspace{-3mm}
\includegraphics[width = .24\linewidth]{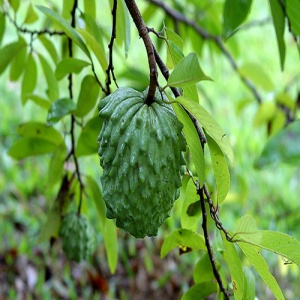}
\hspace{-3mm}
\includegraphics[width = .24\linewidth]{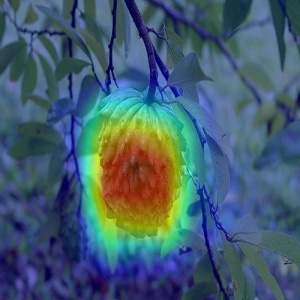}
\\
\vspace{-1mm}
\includegraphics[width = .24\linewidth]{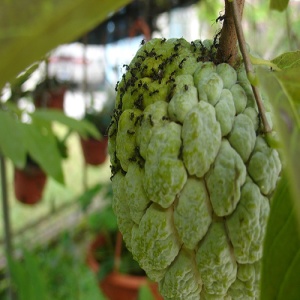}
\hspace{-3mm}
\includegraphics[width = .24\linewidth]{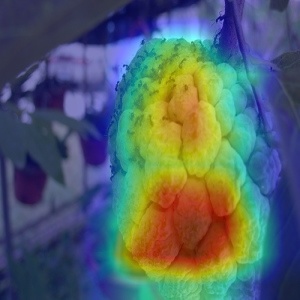}
\hspace{-3mm}
\includegraphics[width = .24\linewidth]{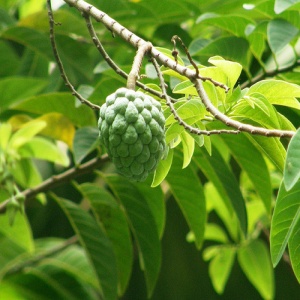}
\hspace{-3mm}
\includegraphics[width = .24\linewidth]{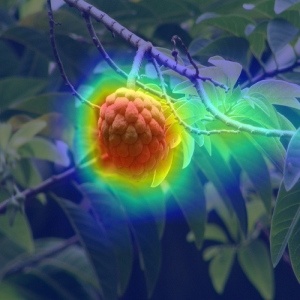}
\caption{Examples of images from the Custard Apple class drawn from the auxiliary dataset. Corresponding heatmaps are also displayed, which represent the regions that most strongly activated the outputs of the target group. These heatmaps were generated by employing the Grad-CAM method.}
\label{fig:gradcam_custard_apple}
\end{figure}

\begin{figure}[!ht]
\centering
\includegraphics[width = .24\linewidth]{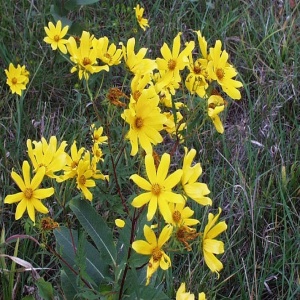}
\hspace{-3mm}
\includegraphics[width = .24\linewidth]{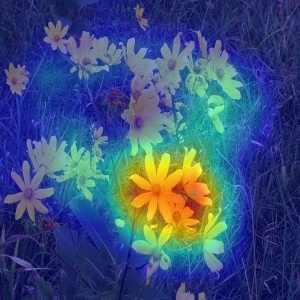}
\hspace{-3mm}
\includegraphics[width = .24\linewidth]{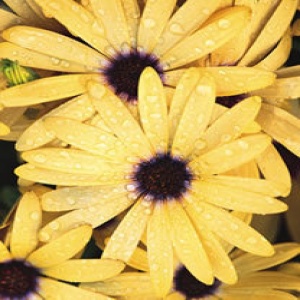}
\hspace{-3mm}
\includegraphics[width = .24\linewidth]{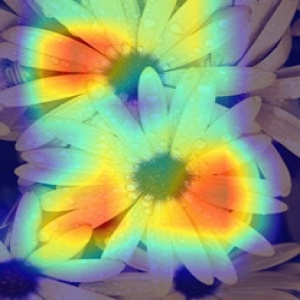}
\\
\vspace{-1mm}
\includegraphics[width = .24\linewidth]{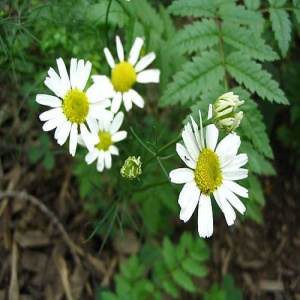}
\hspace{-3mm}
\includegraphics[width = .24\linewidth]{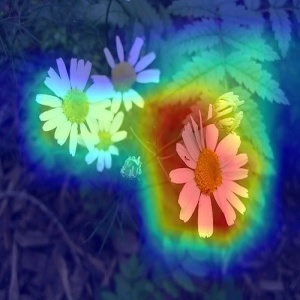}
\hspace{-3mm}
\includegraphics[width = .24\linewidth]{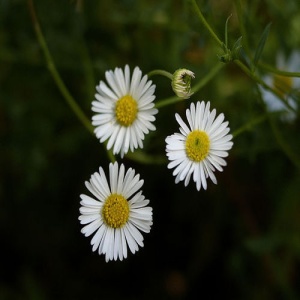}
\hspace{-3mm}
\includegraphics[width = .24\linewidth]{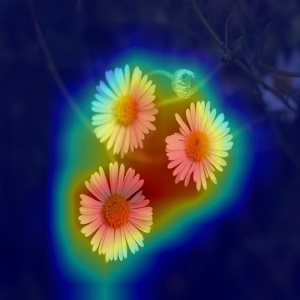}
\caption{Examples of images from the Daisy class drawn from the auxiliary dataset. Corresponding heatmaps are also displayed, which represent the regions that most strongly activated the outputs of the target group. These heatmaps were generated by employing the Grad-CAM method.}
\label{fig:gradcam_daisy}
\end{figure}

\begin{figure}[!ht]
\centering
\includegraphics[width = .24\linewidth]{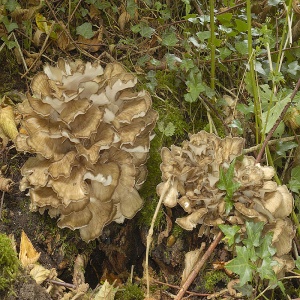}
\hspace{-3mm}
\includegraphics[width = .24\linewidth]{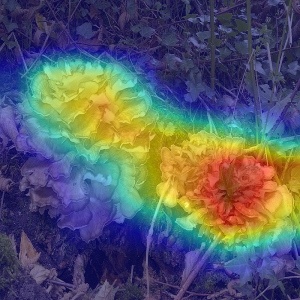}
\hspace{-3mm}
\includegraphics[width = .24\linewidth]{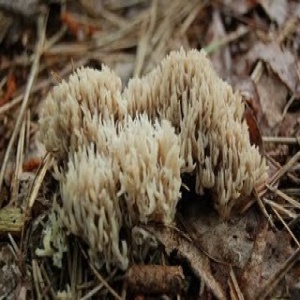}
\hspace{-3mm}
\includegraphics[width = .24\linewidth]{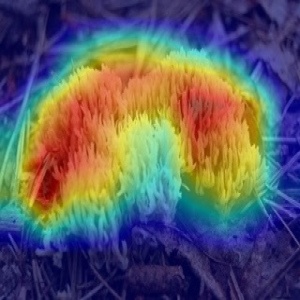}
\\
\vspace{-1mm}
\includegraphics[width = .24\linewidth]{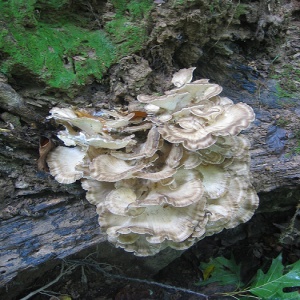}
\hspace{-3mm}
\includegraphics[width = .24\linewidth]{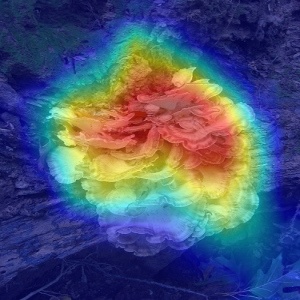}
\hspace{-3mm}
\includegraphics[width = .24\linewidth]{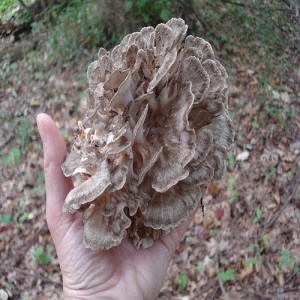}
\hspace{-3mm}
\includegraphics[width = .24\linewidth]{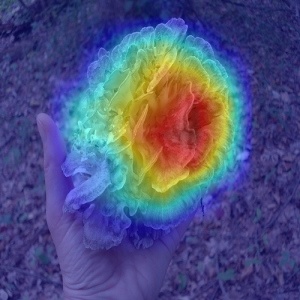}
\caption{Examples of images from the Maitake class drawn from the auxiliary dataset. Corresponding heatmaps are also displayed, which represent the regions that most strongly activated the outputs of the target group. These heatmaps were generated by employing the Grad-CAM method.}
\label{fig:gradcam_maitake}
\end{figure}

\begin{figure}[!ht]
\centering
\includegraphics[width = .24\linewidth]{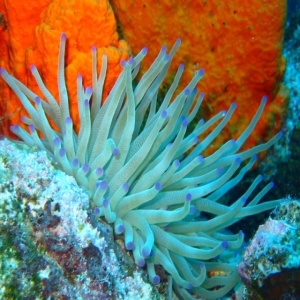}
\hspace{-3mm}
\includegraphics[width = .24\linewidth]{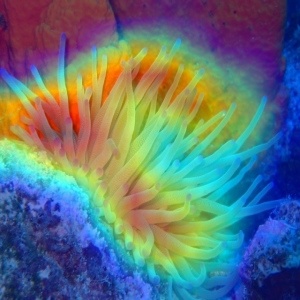}
\hspace{-3mm}
\includegraphics[width = .24\linewidth]{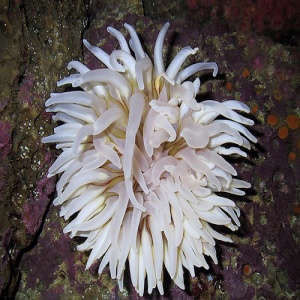}
\hspace{-3mm}
\includegraphics[width = .24\linewidth]{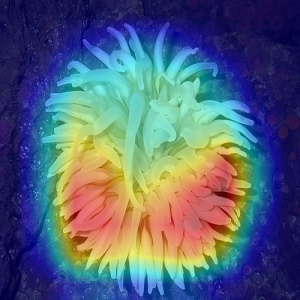}
\\
\vspace{-1mm}
\includegraphics[width = .24\linewidth]{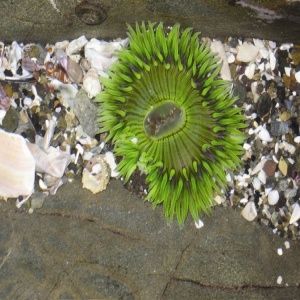}
\hspace{-3mm}
\includegraphics[width = .24\linewidth]{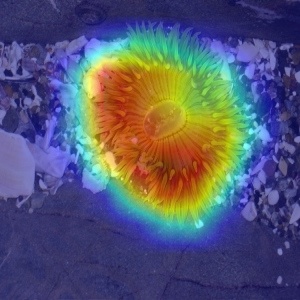}
\hspace{-3mm}
\includegraphics[width = .24\linewidth]{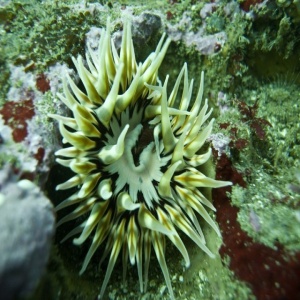}
\hspace{-3mm}
\includegraphics[width = .24\linewidth]{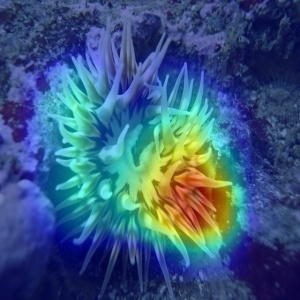}
\caption{Examples of images from the Sea Anemone class drawn from the auxiliary dataset. Corresponding heatmaps are also displayed, which represent the regions that most strongly activated the outputs of the target group. These heatmaps were generated by employing the Grad-CAM method.}
\label{fig:gradcam_sea_anemone}
\end{figure}

To investigate the relationship between the auxiliary dataset and the target dataset, the Grad-CAM technique is applied to the classes of the auxiliary dataset that most significantly activated the model's outputs corresponding to the target group. Section \ref{sec:appendix} illustrates the top 10 classes from each auxiliary dataset with the highest activation. However, this section will focus solely on ImageNet, as PlantNet images predominantly consist of leaves and plants, which trivially relate to the target dataset. Figs. \ref{fig:gradcam_coral_fungus}, \ref{fig:gradcam_custard_apple}, \ref{fig:gradcam_daisy}, \ref{fig:gradcam_maitake}, and \ref{fig:gradcam_sea_anemone} display selected images and their corresponding heatmaps, highlighting the regions that most contributed to the model's interpretation of them as belonging to the target group. These classes are, respectively: coral fungus, custard apple, daisy, maitake mushroom, and sea anemone. A commonality among these classes is that they all represent organic elements. Although ImageNet encompasses numerous images of artificial objects, such as vehicles, equipment, and tools, the model did not associate them with the target dataset. It appears that the model discerned a relationship between these living, organic materials, and hop leaves, despite the absence of shared color, shape, or texture attributes.

\section{Conclusion}
\label{sec:conclusion}

%Nós apresentamos o Simultaneous Learning, uma nova abordagem para melhorar a regularização de modelos de classificação, utilizando bases de dados já existentes que não necessariamente estejam relacionadas ao dataset alvo. A técnica requer apenas a modificação da última camada do modelo e o aumento no número de parâmetros é pequeno, tornando-a escalável. Além disso, tem pouco impacto no treinamento, uma vez que o tamanho do lote é mantido. Propomos ainda uma penalização inter-grupo para auxiliar no treinamento, separando o grupo alvo do grupo auxiliar.

%We introduced Simultaneous Learning, a new approach to improving the regularization of classification models, using existing datasets that are not necessarily related to the target dataset. The technique requires only the modification of the model's last layer and the increase in the number of parameters is small, making it scalable. Moreover, it has little impact on training since the batch size is maintained. We also proposed an inter-group penalty to aid in training, separating the target group from the auxiliary group.

In this study, we present the method of Simultaneous Learning, an innovative approach for enhancing the regularization of classification models by utilizing existing datasets that may not be directly related to the target dataset. This method entails solely modifying the final layer of the model to accommodate the class groups and employing the Simultaneous Learning Loss for the training process. This approach results in a minimal increase in the number of parameters, thereby ensuring scalability. Additionally, it has a negligible impact on training, as the batch size remains constant. The proposed loss function incorporates an inter-group penalty designed to facilitate training by distinguishing between the target and auxiliary groups.

%Os resultados foram significativos, superando até mesmo o dropout, uma técnica de regularização amplamente utilizada. A técnica mostrou-se eficiente em bases de dados muito reduzidas, o que pode ser promissor para uma variedade de problemas semelhantes. Adicionalmente, testamos a técnica em conjunto com o dropout e, ainda assim, houve melhoria na generalização.

%The results were significant, outperforming even dropout, a widely used regularization technique. The method proved efficient on highly reduced datasets, which can be promising for a variety of similar problems. Additionally, we tested the method in combination with dropout, and there was still an improvement in generalization.

The findings were noteworthy, surpassing the performance of dropout—a commonly employed regularization technique—when applied to the UFOP-HVD, our target dataset. The method also demonstrated efficiency with significantly reduced sample sizes, offering promise for a range of similar challenges. Moreover, when tested in conjunction with dropout, the approach still yielded improvements in generalization, achieving state-of-the-art results for the UFOP-HVD dataset. 

Furthermore, we introduced a novel technique called Layer Correlation to compare the quality of features generated by two models in a layer-by-layer manner. This approach was employed to compare models with and without the Simultaneous Learning method, leading to significant findings. Of paramount importance is the versatility of Layer Correlation; it is not confined to the studied models and presents the potential for broader application across a variety of models featuring convolutional layers. Additionally, we tested the Grad-CAM method for visualizing how the auxiliary dataset triggers the outputs of the target group, providing a complementary perspective. Ultimately, both Layer Correlation and Grad-CAM were leveraged to provide greater interpretability to our results and process.

%Também apresentamos um método para comparar a qualidade das features geradas por dois modelos. O método não está restrito à nossa técnica e pode ser aplicado a qualquer modelo com redes neurais convolucionais.

%We also presented a method for comparing the quality of features generated by two models. The method is not restricted to our technique and can be applied to any model with convolutional neural networks.

%Como trabalhos futuros, pretendemos avaliar o Simultaneous Learning em outras bases de dados, tanto como alvo quanto como auxiliar. Além disso, como foi utilizado apenas um subconjunto das bases de dados auxiliares, seria interessante testar com o conjunto completo e analisar o impacto. Comparar com mais técnicas de regularização pode atestar a robustez do método. Por fim, estender o modelo para outros tipos de rede, como regressão e detecção de objetos, é um importante objeto de estudo.

%In future work, we intend to evaluate Simultaneous Learning on other datasets, both as targets and as auxiliary. Furthermore, since only a subset of the auxiliary datasets was used, it would be interesting to test with the complete set and analyze the impact. Comparing with more regularization techniques can attest to the method's robustness. Finally, extending the model to other types of networks, such as regression and object detection, is an important area of study.

Within the scope of this research, we acknowledge certain limitations. Firstly, the hyperparameters $\lambda$ related to the weight of each group and the inter-group penalties are manually adjusted, suggesting that further investigation into methods for automatically optimizing these values could be beneficial. Secondly, while the proposed model is well-suited for single-class classification problems, it may require reevaluation for multi-label issues where a single instance can contain more than one class. In addition, the method currently utilizes only one auxiliary database per training; future research could explore using multiple auxiliary databases to investigate their impact on performance.

In future research, the method of Simultaneous Learning could be evaluated across various datasets and contexts. Additionally, considering that only a subset of the auxiliary datasets was utilized, it would be intriguing to assess the method with more data and analyze the impact. Lastly, extending the model to encompass other types of networks, such as regression and object detection, could present an interesting path for investigation.

\section*{Declaration of Competing Interest}

The authors declare that they have no known competing financial interests or personal relationships that could have appeared to influence the work reported in this paper. 

\section*{Acknowledgments}

The authors would also like to thank the \textit{Coordenação de Aperfeiçoamento de Pessoal de Nível Superior} - Brazil (CAPES) - Finance Code 001, \textit{Fundacão de Amparo à Pesquisa do Estado de Minas Gerais} (FAPEMIG, grants APQ-01518-21), \textit{Conselho Nacional de Desenvolvimento Científico e Tecnológico} (CNPq, grants 308400/2022-4) and Universidade Federal de Ouro Preto (PROPPI/UFOP) for supporting the development of this study.

\newpage
\bibliography{elsarticle-template}

\begin{thebibliography}{10}
\expandafter\ifx\csname url\endcsname\relax
  \def\url#1{\texttt{#1}}\fi
\expandafter\ifx\csname urlprefix\endcsname\relax\def\urlprefix{URL }\fi
\expandafter\ifx\csname href\endcsname\relax
  \def\href#1#2{#2} \def\path#1{#1}\fi

\bibitem{lecun2015deep}
Y.~LeCun, Y.~Bengio, G.~Hinton, Deep learning, nature 521~(7553) (2015)
  436--444.

\bibitem{reedha2022transformer}
R.~Reedha, E.~Dericquebourg, R.~Canals, A.~Hafiane, Transformer neural network
  for weed and crop classification of high resolution uav images, Remote
  Sensing 14~(3) (2022) 592.

\bibitem{makanapura2022classification}
N.~Makanapura, C.~Sujatha, P.~R. Patil, P.~Desai, Classification of plant
  seedlings using deep convolutional neural network architectures, in: Journal
  of Physics: Conference Series, Vol. 2161, IOP Publishing, 2022, p. 012006.

\bibitem{wan2022plant}
W.~M. A. H.~B. Wan, S.~Nordin, et~al., Plant recognition system using
  convolutional neural network, in: IOP Conference Series: Earth and
  Environmental Science, Vol. 1019, IOP Publishing, 2022, p. 012031.

\bibitem{domingos2012few}
P.~Domingos, A few useful things to know about machine learning, Communications
  of the ACM 55~(10) (2012) 78--87.

\bibitem{tian2022comprehensive}
Y.~Tian, Y.~Zhang, A comprehensive survey on regularization strategies in
  machine learning, Information Fusion 80 (2022) 146--166.

\bibitem{goodfellow2016deep}
I.~Goodfellow, Y.~Bengio, A.~Courville, Deep Learning, MIT Press, 2016,
  \url{http://www.deeplearningbook.org}.

\bibitem{kukavcka2017regularization}
J.~Kuka{\v{c}}ka, V.~Golkov, D.~Cremers, Regularization for deep learning: A
  taxonomy, arXiv preprint arXiv:1710.10686 (2017).

\bibitem{krogh1991simple}
A.~Krogh, J.~Hertz, A simple weight decay can improve generalization, Advances
  in Neural Information Processing Systems 4 (1991).

\bibitem{srivastava2014dropout}
N.~Srivastava, G.~Hinton, A.~Krizhevsky, I.~Sutskever, R.~Salakhutdinov,
  Dropout: a simple way to prevent neural networks from overfitting, The
  Journal of Machine Learning Research 15~(1) (2014) 1929--1958.

\bibitem{warde2013empirical}
D.~Warde-Farley, I.~J. Goodfellow, A.~Courville, Y.~Bengio, An empirical
  analysis of dropout in piecewise linear networks, arXiv preprint
  arXiv:1312.6197 (2013).

\bibitem{ioffe2015batch}
S.~Ioffe, C.~Szegedy, Batch normalization: Accelerating deep network training
  by reducing internal covariate shift, in: International Conference on Machine
  Learning, pmlr, 2015, pp. 448--456.

\bibitem{shorten2019survey}
C.~Shorten, T.~M. Khoshgoftaar, A survey on image data augmentation for deep
  learning, Journal of Big Data 6~(1) (2019) 1--48.

\bibitem{zhang2005boosting}
T.~Zhang, B.~Yu, Boosting with early stopping: Convergence and consistency, The
  Annals of Statistics 33 (2005).

\bibitem{wong2020fast}
E.~Wong, L.~Rice, J.~Z. Kolter, Fast is better than free: Revisiting
  adversarial training, in: International Conference on Learning
  Representations, 2020, pp. 1--17.

\bibitem{pan2010survey}
S.~J. Pan, Q.~Yang, A survey on transfer learning, IEEE Transactions on
  Knowledge and Data Engineering 22~(10) (2010) 1345--1359.

\bibitem{caruana1998multitask}
R.~Caruana, Multitask learning, Springer, 1998.

\bibitem{moradi2020survey}
R.~Moradi, R.~Berangi, B.~Minaei, A survey of regularization strategies for
  deep models, Artificial Intelligence Review 53 (2020) 3947--3986.

\bibitem{ruder2017overview}
S.~Ruder, An overview of multi-task learning in deep neural networks, arXiv
  preprint arXiv:1706.05098 (2017).

\bibitem{abbas2023secure}
T.~Abbas, A.~Fatima, T.~Shahzad, K.~Alissa, T.~M. Ghazal, M.~M. Al-Sakhnini,
  S.~Abbas, M.~A. Khan, A.~Ahmed, et~al., Secure iomt for disease prediction
  empowered with transfer learning in healthcare 5.0, the concept and case
  study, IEEE Access (2023).

\bibitem{huang2018automatic}
H.~Huang, H.~Deng, J.~Chen, L.~Han, W.~Wang, Automatic multi-task learning
  system for abnormal network traffic detection., International Journal of
  Emerging Technologies in Learning 13~(4) (2018).

\bibitem{wang2020environment}
X.~E. Wang, V.~Jain, E.~Ie, W.~Y. Wang, Z.~Kozareva, S.~Ravi,
  Environment-agnostic multitask learning for natural language grounded
  navigation, in: Computer Vision--ECCV 2020: 16th European Conference,
  Glasgow, UK, August 23--28, 2020, Proceedings, Part XXIV 16, Springer, 2020,
  pp. 413--430.

\bibitem{li2021multi}
Y.~Li, C.~Caragea, A multi-task learning framework for multi-target stance
  detection, in: Findings of the Association for Computational Linguistics:
  ACL-IJCNLP 2021, 2021, pp. 2320--2326.

\bibitem{kumar2012learning}
A.~Kumar, H.~Daume~III, Learning task grouping and overlap in multi-task
  learning, arXiv preprint arXiv:1206.6417 (2012).

\bibitem{zhao2018modulation}
X.~Zhao, H.~Li, X.~Shen, X.~Liang, Y.~Wu, A modulation module for multi-task
  learning with applications in image retrieval, in: Proceedings of the
  European Conference on Computer Vision (ECCV), 2018, pp. 401--416.

\bibitem{zhang2018overview}
Y.~Zhang, Q.~Yang, An overview of multi-task learning, National Science Review
  5~(1) (2018) 30--43.

\bibitem{zhang2021survey}
Y.~Zhang, Q.~Yang, A survey on multi-task learning, IEEE Transactions on
  Knowledge and Data Engineering 34~(12) (2021) 5586--5609.

\bibitem{deng2009imagenet}
J.~Deng, W.~Dong, R.~Socher, L.-J. Li, K.~Li, L.~Fei-Fei, Imagenet: A
  large-scale hierarchical image database, in: 2009 IEEE Conference on Computer
  Vision and Pattern Recognition, Ieee, 2009, pp. 248--255.

\bibitem{russakovsky2015imagenet}
O.~Russakovsky, J.~Deng, H.~Su, J.~Krause, S.~Satheesh, S.~Ma, Z.~Huang,
  A.~Karpathy, A.~Khosla, M.~Bernstein, et~al., Imagenet large scale visual
  recognition challenge, International Journal of Computer Vision 115 (2015)
  211--252.

\bibitem{burgos2010analysis}
X.~P. Burgos-Artizzu, A.~Ribeiro, A.~Tellaeche, G.~Pajares,
  C.~Fern{\'a}ndez-Quintanilla, Analysis of natural images processing for the
  extraction of agricultural elements, Image and Vision Computing 28~(1) (2010)
  138--149.

\bibitem{hameed2018comprehensive}
K.~Hameed, D.~Chai, A.~Rassau, A comprehensive review of fruit and vegetable
  classification techniques, Image and Vision Computing 80 (2018) 24--44.

\bibitem{lee2017automatic}
H.-H. Lee, K.-S. Hong, Automatic recognition of flower species in the natural
  environment, Image and Vision Computing 61 (2017) 98--114.

\bibitem{healey}
J.~Healey, The Hops List: 265 Beer Hop Varieties From Around the World, Julian
  Healey, 2016.

\bibitem{jenks2011plant}
M.~A. Jenks, Plant nomenclature, Purdue University-Department of Horticulture
  and Landscape Architecture. Dispon{\'\i}vel (2011).

\bibitem{garcin2021plantnet}
C.~Garcin, A.~Joly, P.~Bonnet, M.~Servajean, J.~Salmon,
  \href{https://doi.org/10.5281/zenodo.4726653}{Pl@ntnet-300k image dataset}
  (Apr. 2021).
\newblock \href {https://doi.org/10.5281/zenodo.4726653}
  {\path{doi:10.5281/zenodo.4726653}}.
\newline\urlprefix\url{https://doi.org/10.5281/zenodo.4726653}

\bibitem{castro2021end}
P.~H.~N. CASTRO, G.~J.~P. Moreira, E.~J. da~Silva~Luz, An end-to-end deep
  learning system for hop classification, IEEE Latin America Transactions
  20~(3) (2021) 430--442.

\bibitem{paredes2012exploiting}
B.~R. Paredes, A.~Argyriou, N.~Berthouze, M.~Pontil, Exploiting unrelated tasks
  in multi-task learning, in: Artificial Intelligence and Statistics, PMLR,
  2012, pp. 951--959.

\bibitem{liebel2018auxiliary}
L.~Liebel, M.~K{\"o}rner, Auxiliary tasks in multi-task learning, arXiv
  preprint arXiv:1805.06334 (2018).

\bibitem{chennupati2019auxnet}
S.~Chennupati, G.~Sistu, S.~Yogamani, S.~Rawashdeh, Auxnet: Auxiliary tasks
  enhanced semantic segmentation for automated driving, arXiv preprint
  arXiv:1901.05808 (2019).

\bibitem{zhu2019ta}
Y.~Zhu, W.~Sun, X.~Cao, C.~Wang, D.~Wu, Y.~Yang, N.~Ye, Ta-cnn: Two-way
  attention models in deep convolutional neural network for plant recognition,
  Neurocomputing 365 (2019) 191--200.

\bibitem{lee2021conditional}
S.~H. Lee, H.~Go{\"e}au, P.~Bonnet, A.~Joly, Conditional multi-task learning
  for plant disease identification, in: 2020 25th international conference on
  pattern recognition (ICPR), IEEE, 2021, pp. 3320--3327.

\bibitem{keceli2022deep}
A.~S. Keceli, A.~Kaya, C.~Catal, B.~Tekinerdogan, Deep learning-based
  multi-task prediction system for plant disease and species detection,
  Ecological Informatics 69 (2022) 101679.

\bibitem{wang2022dhbp}
D.~Wang, J.~Wang, Z.~Ren, W.~Li, Dhbp: A dual-stream hierarchical bilinear
  pooling model for plant disease multi-task classification, Computers and
  Electronics in Agriculture 195 (2022) 106788.

\bibitem{zhuang2020comprehensive}
F.~Zhuang, Z.~Qi, K.~Duan, D.~Xi, Y.~Zhu, H.~Zhu, H.~Xiong, Q.~He, A
  comprehensive survey on transfer learning, Proceedings of the IEEE 109~(1)
  (2020) 43--76.

\bibitem{kaya2019analysis}
A.~Kaya, A.~S. Keceli, C.~Catal, H.~Y. Yalic, H.~Temucin, B.~Tekinerdogan,
  Analysis of transfer learning for deep neural network based plant
  classification models, Computers and Electronics in Agriculture 158 (2019)
  20--29.

\bibitem{espejo2020improving}
B.~Espejo-Garcia, N.~Mylonas, L.~Athanasakos, S.~Fountas, Improving weeds
  identification with a repository of agricultural pre-trained deep neural
  networks, Computers and Electronics in Agriculture 175 (2020) 105593.

\bibitem{espejo2020towards}
B.~Espejo-Garcia, N.~Mylonas, L.~Athanasakos, S.~Fountas, I.~Vasilakoglou,
  Towards weeds identification assistance through transfer learning, Computers
  and Electronics in Agriculture 171 (2020) 105306.

\bibitem{ahmad2021performance}
A.~Ahmad, D.~Saraswat, V.~Aggarwal, A.~Etienne, B.~Hancock, Performance of deep
  learning models for classifying and detecting common weeds in corn and
  soybean production systems, Computers and Electronics in Agriculture 184
  (2021) 106081.

\bibitem{pratondo2022classification}
A.~Pratondo, E.~Elfahmi, A.~Novianty, Classification of curcuma longa and
  curcuma zanthorrhiza using transfer learning, PeerJ Computer Science 8 (2022)
  e1168.

\bibitem{chen2022performance}
D.~Chen, Y.~Lu, Z.~Li, S.~Young, Performance evaluation of deep transfer
  learning on multi-class identification of common weed species in cotton
  production systems, Computers and Electronics in Agriculture 198 (2022)
  107091.

\bibitem{wan2013regularization}
L.~Wan, M.~Zeiler, S.~Zhang, Y.~Le~Cun, R.~Fergus, Regularization of neural
  networks using dropconnect, in: International Conference on Machine Learning,
  PMLR, 2013, pp. 1058--1066.

\bibitem{ba2013adaptive}
J.~Ba, B.~Frey, Adaptive dropout for training deep neural networks, Advances in
  Neural Information Processing Systems 26 (2013).

\bibitem{morerio2017curriculum}
P.~Morerio, J.~Cavazza, R.~Volpi, R.~Vidal, V.~Murino, Curriculum dropout, in:
  Proceedings of the IEEE International Conference on Computer Vision, 2017,
  pp. 3544--3552.

\bibitem{moradi2019sparsemaps}
R.~Moradi, R.~Berangi, B.~Minaei, Sparsemaps: convolutional networks with
  sparse feature maps for tiny image classification, Expert Systems with
  Applications 119 (2019) 142--154.

\bibitem{pham2021autodropout}
H.~Pham, Q.~Le, Autodropout: Learning dropout patterns to regularize deep
  networks, in: Proceedings of the AAAI Conference on Artificial Intelligence,
  Vol.~35, 2021, pp. 9351--9359.

\bibitem{lu2021localdrop}
Z.~Lu, C.~Xu, B.~Du, T.~Ishida, L.~Zhang, M.~Sugiyama, Localdrop: A hybrid
  regularization for deep neural networks, IEEE Transactions on Pattern
  Analysis and Machine Intelligence 44~(7) (2021) 3590--3601.

\bibitem{liu2016flower}
Y.~Liu, F.~Tang, D.~Zhou, Y.~Meng, W.~Dong, Flower classification via
  convolutional neural network, in: 2016 IEEE International Conference on
  Functional-Structural Plant Growth Modeling, Simulation, Visualization and
  Applications (FSPMA), IEEE, 2016, pp. 110--116.

\bibitem{akter2020cnn}
R.~Akter, M.~I. Hosen, Cnn-based leaf image classification for bangladeshi
  medicinal plant recognition, in: 2020 Emerging Technology in Computing,
  Communication and Electronics (ETCCE), IEEE, 2020, pp. 1--6.

\bibitem{wang2020fruit}
S.-H. Wang, Y.~Chen, Fruit category classification via an eight-layer
  convolutional neural network with parametric rectified linear unit and
  dropout technique, Multimedia Tools and Applications 79 (2020) 15117--15133.

\bibitem{haichen2020weeds}
J.~Haichen, C.~Qingrui, L.~Zheng~Guang, Weeds and crops classification using
  deep convolutional neural network, in: Proceedings of the 3rd International
  Conference on Control and Computer Vision, 2020, pp. 40--44.

\bibitem{chauhan2021deep}
P.~Chauhan, H.~L. Mandoria, A.~Negi, Deep residual neural network for plant
  seedling image classification, Agricultural Informatics: Automation Using the
  IoT and Machine Learning (2021) 131--146.

\bibitem{van2017l2}
T.~Van~Laarhoven, L2 regularization versus batch and weight normalization,
  arXiv preprint arXiv:1706.05350 (2017).

\bibitem{he2019bag}
T.~He, Z.~Zhang, H.~Zhang, Z.~Zhang, J.~Xie, M.~Li, Bag of tricks for image
  classification with convolutional neural networks, in: Proceedings of the
  IEEE/CVF Conference on Computer Vision and Pattern Recognition, 2019, pp.
  558--567.

\bibitem{li2019understanding}
X.~Li, S.~Chen, X.~Hu, J.~Yang, Understanding the disharmony between dropout
  and batch normalization by variance shift, in: Proceedings of the IEEE/CVF
  Conference on Computer Vision and Pattern Recognition, 2019, pp. 2682--2690.

\bibitem{bhatnagar2017classification}
S.~Bhatnagar, D.~Ghosal, M.~H. Kolekar, Classification of fashion article
  images using convolutional neural networks, in: 2017 Fourth International
  Conference on Image Information Processing (ICIIP), IEEE, 2017, pp. 1--6.

\bibitem{ho2019real}
Y.~Ho, S.~Wookey, The real-world-weight cross-entropy loss function: Modeling
  the costs of mislabeling, IEEE Access 8 (2019) 4806--4813.

\bibitem{lin2013network}
M.~Lin, Q.~Chen, S.~Yan, Network in network, arXiv preprint arXiv:1312.4400
  (2013).

\bibitem{rodgers1988thirteen}
J.~L. Rodgers, W.~A. Nicewander, Thirteen ways to look at the correlation
  coefficient, American statistician (1988) 59--66.

\bibitem{chandrashekar2014survey}
G.~Chandrashekar, F.~Sahin, A survey on feature selection methods, Computers \&
  Electrical Engineering 40~(1) (2014) 16--28.

\bibitem{selvaraju2017grad}
R.~R. Selvaraju, M.~Cogswell, A.~Das, R.~Vedantam, D.~Parikh, D.~Batra,
  Grad-cam: Visual explanations from deep networks via gradient-based
  localization, in: Proceedings of the IEEE International Conference on
  Computer Vision, 2017, pp. 618--626.

\bibitem{szegedy2016rethinking}
C.~Szegedy, V.~Vanhoucke, S.~Ioffe, J.~Shlens, Z.~Wojna, Rethinking the
  inception architecture for computer vision, in: Proceedings of the IEEE
  Conference on Computer Vision and Pattern Recognition, 2016, pp. 2818--2826.

\bibitem{he2016deep}
K.~He, X.~Zhang, S.~Ren, J.~Sun, Deep residual learning for image recognition,
  in: Proceedings of the IEEE Conference on Computer Vision and Pattern
  Recognition, 2016, pp. 770--778.

\bibitem{duchi2011adaptive}
J.~Duchi, E.~Hazan, Y.~Singer, Adaptive subgradient methods for online learning
  and stochastic optimization., Journal of Machine Learning Research 12~(7)
  (2011).

\bibitem{glorot2010understanding}
X.~Glorot, Y.~Bengio, Understanding the difficulty of training deep feedforward
  neural networks, in: Proceedings of the thirteenth international conference
  on artificial intelligence and statistics, JMLR Workshop and Conference
  Proceedings, 2010, pp. 249--256.

\bibitem{castro2021dataset}
P.~Castro, E.~Luz, G.~Moreira, Dataset for hop varieties classification, Data
  in Brief 38 (2021) 107312.

\end{thebibliography}

\newpage
\appendix
\section{Auxiliary classes with the highest activation of the target group.}
\label{sec:appendix}

\begin{figure}[!ht]
\centering
\includegraphics[width = .98\linewidth]{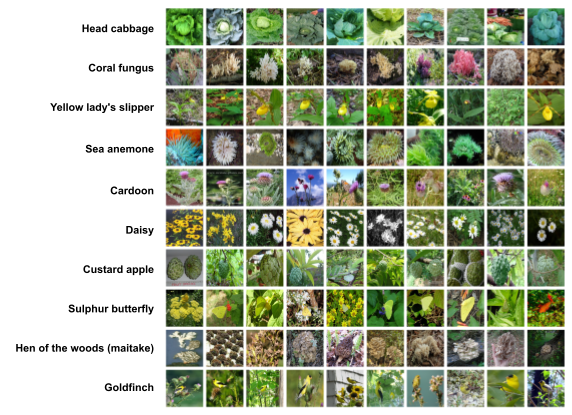} 
\caption{Top 10 ImageNet classes that activated the target group the most. Each row corresponds to a class and its 10 instances with the highest activation.}
\label{fig:most_activated_instance_per_class_imagenet}
\end{figure}

\begin{figure}[!ht]
\centering
\includegraphics[width = .98\linewidth]{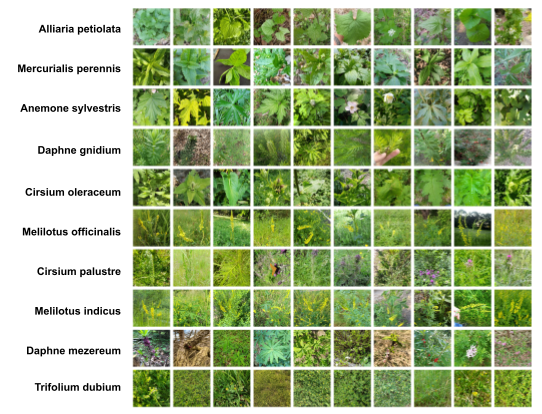} 
\caption{Top 10 PlantNet classes that activated the target group the most. Each row corresponds to a class and its 10 instances with the highest activation.}
\label{fig:most_activated_instance_per_class_plantnet}
\end{figure}

\end{document}